\documentclass{article}
\usepackage[preprint]{neurips_2025}
\usepackage[utf8]{inputenc} 
\usepackage[T1]{fontenc}    
\usepackage[backref=page,colorlinks,citecolor=violet]{hyperref}       
\usepackage{url}            
\usepackage{booktabs}       
\usepackage{amsfonts}       
\usepackage{nicefrac}       
\usepackage{microtype}      
\usepackage{xcolor}  
\usepackage{graphicx}
\usepackage{natbib}

\usepackage{pifont}
\usepackage{adjustbox}
\usepackage{verbatim}
\usepackage{amsfonts}
\usepackage{amsmath}
\usepackage{amsthm}
\usepackage{algorithm}
\usepackage{algorithmic}
\usepackage{mathtools}
\usepackage{bbold}
\usepackage{thmtools} 
\usepackage{thm-restate}
\usepackage{enumerate}
\usepackage{float}
\usepackage{enumitem}
\usepackage{cancel}
\usepackage{amsmath,amssymb, amsfonts}
\usepackage{color}

\usepackage{pgfplots}
\usetikzlibrary{external}
\usepackage{glossaries}
\usepackage{geometry}
\usepackage{longtable}
\usepackage{etoolbox}
\usepackage[refpage,nomentbl,stdsubgroups]{nomencl}
\setnomtableformat{lllll}

\usepackage{xcolor}
\usepgfplotslibrary{groupplots}
\usetikzlibrary{arrows.meta}

\renewcommand\nomgroup[1]{%
  \item[\bfseries
  \ifstrequal{#1}{P}{Problem Parameters}{%
  \ifstrequal{#1}{M}{Model Details}{%
  \ifstrequal{#1}{A}{Algorithmic Details}{}}}%
]}

\renewcommand{\nomgroup}[1]{
\item[]\hspace*{-\leftmargin}%
\rule[2pt]{0.45\linewidth}{1pt}%
\hfill #1\hfill \rule[2pt]{0.45\linewidth}{1pt}}

\newtheorem{definition}{Definition}
\newtheorem{theorem}{Theorem}
\newtheorem{lemma}{Lemma}

\newtheorem{assumption}{Assumption}
\newtheorem{proposition}{Proposition}

\newtheorem{exmp}{Example}
\newtheorem{remark}{Remark}[section]

\usepackage{mathtools}
\usepackage{bbold}
\usepackage{soul}
\usepackage{subcaption}
\usepackage{wrapfig}
\input{shared/boondox-cal.sty}
\usepackage[nohints]{minitoc}

\usepackage{microtype}
\usepackage{hyperref}
\def\mE{\mathbb{E}}

\usepackage{color}

\def\cT{\mathcal{T}}
\def\cE{\mathcal{E}}

\def\cX{\mathcal{X}}
\def\cF{\mathcal{F}}
\def\cR{\mathcal{R}}

\def\mR{\mathbb{R}}

\def\cN{\mathcal{N}}

\def\cA{\mathcal{A}}

\def\cP{\mathcal{P}}
\def\cM{\mathcal{M}}

\def\cG{\mathcal{G}}
\def\cG{\mathcal{G}}
\def\cB{\mathcal{B}}
\def\cH{\mathcal{H}}
\def\cZ{\mathcal{Z}}
\def\cL{\mathcal{L}}

\def\mN{\mathbb{N}}
\def\mN{\mathbb{N}}

\def\mZ{\mathbb{Z}}

\def\cspace{\cX}
\def\aspace{[K]}

\def\numobj{M}

\def\weakdom{\preccurlyeq}

\def\notdom{ \ \cancel{\prec}_{\cone}\ }

\def\tree{\cT}
\def\binrad{V_h}

\def\leaft{\cL_t}
\def\gevent{\cG}

\def\cC{\mathcal{C}}
\def\paretospace{\mathcal{Z}}
\def\pftrue{\mathcal{P}^{\ast}}
\def\leaf{\mathcal{L}}

\def\numchild{\Psi}

\def\armset{[K]}
\def\vectle{\preceq_{\cC}} 
\def\vectl{\prec_{\cC}}

\def\uncertrad{\bar{u}_{k,t}(h,i)}
\def\horizon{T}
\def\cone{\mathcal{C}}
\def\pfestm{\hat{\cP}(X_t)}
\def\leaft{\mathcal{L}_{t}}

\def\activearm{\mathcal{A}_{(h_t,i_t)}}
\def\currbin{(h_t,i_t)}
\def\regt{R(T)}
\def\treepart{\cT}
\def\pspace{\cZ}
\def\binid{(h,i)}

\def\source{P}
\def\target{Q}

\newcommand{\ind}[1]{\mathbb{1}\left(#1\right)}

\newcommand{\upktl}[3]{U^{(#3)}_{#1,#2}}
\newcommand{\lowktl}[3]{L^{(#3)}_{#1,#2}}
\newcommand{\bin}[2]{\mathcal{B}_{(#1,#2)}}
\newcommand{\pdist}[2]{d_{p}\left(#1,#2\right)}
\newcommand{\muestm}[3]{\hat{\mu}^{(#1)}_{#2,#3}}
\newcommand{\mutrue}[2]{\mu^{(#1)}_{#2}}

\newcommand{\anc}[1]{\text{anc}(#1)}
\newcommand{\numkt}[1]{n_{k,t}#1}
\newcommand{\pf}[1]{\cP(#1)}

\newcommand{\size}[1]{\vert #1 \vert}

\title{Vector preference-based contextual bandits under distributional shifts}
\author{%
 Apurv Shukla \\
 apurv.shukla@umich.edu\\
 Department of EECS, University of Michigan, Ann Arbor
 \And
 P.R. Kumar \\
 prk@tamu.edu \\
 Department of ECE, Texas A\&M University, College Station
}

\begin{document}
\maketitle

\begin{abstract}
We consider contextual bandit learning under distribution shift when reward vectors are ordered according to a given preference cone. We propose an adaptive-discretization and optimistic elimination based policy that self-tunes to the underlying distribution shift. To measure the performance of this policy, we introduce the notion of preference-based regret which measures the performance of a policy in terms of distance between Pareto fronts. We study the performance of this policy by establishing upper bounds on its regret under various assumptions on the nature of distribution shift. Our regret bounds generalize known results for the existing case of no distribution shift and vectorial reward settings, and scale gracefully with problem parameters in presence of distribution shifts.
\end{abstract}

\section{Introduction}
\label{sec:introduction}
\subsection{Background and Motivation}
Phase I clinical trials are designed to determine the optimal dosage level of newly introduced drugs for further clinical investigation, and therefore, require simultaneously estimating several properties associated with a drug such as its safety, efficacy, and potency, leading to observations in the form of a reward vector rather than a scalar (see~\cite{liang2009multiple}). Typically, these objectives are learned independently by modeling them as contextual bandit problems. In a contextual bandit problem, the decision maker sequentially observes the contexts associated with every incoming patient and prescribes a (context-dependent) dosage. An often overlooked issue is the evolution of context distribution during the duration of the trial. A trial typically lasts several months during which the distribution from which the patient's contexts are sampled changes. Other examples where preference-based learning under distribution shift plays a key role include fair regression, multi-task and meta-learning. The goal in these problems is to learn several competing objectives, ranked according to a preference, (specified through a cone) from training samples (source distribution) and evaluate performance on test distribution or new unseen tasks (target distribution). 

\begin{exmp}
\label{eg:clinical-trial}
To motivate the need for a distribution shift in this setting, consider a clinical trial where the decision maker aims to find the safety-efficacy curve of a given drug subject to a continuous dosage level~\cite{klarner2023drug}. A patient arrives for a clinical trial with covariates describing their demographics, which can change with time, and the clinic responds with drug dosage levels specific to the patient's covariates. The patient covariates thus map the drug dose levels to a feature vector and the expected reward is an unknown nonparametric function of the covariates. Due to changing demographics, the context distribution changes with time, and an efficient policy is needed to learn the safety-efficacy curve having minimal information about it.
\end{exmp}

Motivated by this, we consider a contextual bandit problem with distribution shifts when vector rewards are $M$-dimensional and ranked according to preferences. While it is possible to solve (in parallel) instances of individual (component-wise) learning problems, such approaches will not identify the set of Pareto optimal solutions because the Pareto optimal solutions are not necessarily optimal for any particular learning problem. Another approach to solving the multi-objective problem would be to scalarize the reward using an appropriate weight vector. However, determination of the optimal scalarization to identify points on the Pareto front remains a challenge. Also, existing contextual bandit models do not account for shift in context distribution. This work aims to fill this gap by considering vector-valued rewards and distribution shifts in a contextual bandit setting.
\subsection{Contributions}
We briefly summarize relevant strands of literature and our contribution to them below.
\begin{enumerate}
\item \textbf{Non-parameteric contextual bandits:} A comprehensive overview of results in multi-armed bandit problems can be found in~\cite{lattimore2020bandit}. In this paper, we consider the multi-armed bandit problem where for each arm, the expected reward is a non-parametric function of the observed context (see~\eqref{eqn:rew-vec}). Non-parametric reward models have much larger representational capacity particularly suited for applications of interest alluded to earlier. Consequently, learning such functions and providing tight performance analysis is much more challenging than in the setting in which we have rewards distributions with finite-dimensional parameters. Such models have been previously considered in a scalar, fixed context distribution setting by~\cite{yang2002randomized,perchet2013multi,rigollet2010nonparametric,hu2020smooth}.  Our work extends this line of work along two directions: we consider vector-valued mean-rewards as opposed to scalar rewards, and we assume that contexts arrive from a time-varying distribution as opposed to adversarial (worst-case arrival, as in~\cite{slivkins2011contextual}) or stochastic (i.i.d fixed-distribution arrival as in~\cite{perchet2013multi}). 

\item \textbf{Learning with vectorial rewards:} Learning with vector-valued rewards has been mostly studied in the case of finite arms without contexts by~\cite{yahyaa2014annealing},~\cite{yahyaa2014knowledge} and~\cite{drugan2013designing}.~\cite{turugay2018multi} study a problem similar to the one considered in this paper, a multi-objective bandit problem with non-parametric mean rewards with adversarial context arrival and a continuum of arms. Along this particular line of work, in Example~\ref{eg:insuff-hausdroff} we first show that the space of Pareto fronts (Definition~\ref{defn:space-pareto-fronts}) defined by the order induced by the preference cone on $[0,1]^{M}$ is not complete under the metric induced by gap-measures between mean rewards considered in previous work~\cite{turugay2018multi,auer2016pareto,kone2023adaptive}. We then propose a metric under which this space is complete and analyse regret using this metric. Our results depend on an appropriate notion of Margin (Assumption~\ref{assmpt:margin}) from past work on classification and scalar valued non-parametric contextual bandit models~\cite{perchet2013multi,audibert2007fast} adapted to the current setting of vector-valued rewards. 

\item \textbf{Distribution Shift:}
Distribution shift has been primarily studied in a classification setting wherein the marginal distribution of the contexts is different between source and target distributions. Policies for this class of problems have been designed based on importance-sampling and distributionally-robust learning~\cite{shimodaira2000improving,ben2007analysis,duchi2019distributionally}. Several other works such as~\cite{singh2021learning} (and the references therein) consider the problem of learning under distribution shift in the full-information setting. For the bandit setting~\cite{cai2024transfer,shukla2022dissertation,suk2020self} consider the non-parametric contextual bandit problem under the distribution shift model proposed by~\cite{kpotufe2018marginal}. Closest to this paper, is the work by~\cite{suk2020self} wherein the authors study a scalar finite-armed non-parametric contextual-bandit problem under the distribution-shift model of~\cite{kpotufe2018marginal}. In this paper, we consider a vectorial finite-armed non-parametric contextual bandit problem under the distribution-shift model of~\cite{pathak2022new}. This generalizes the model for distribution-shift and extends it to the setting with vector rewards.
\end{enumerate}

Further related literature is surveyed in Appendix~\ref{sec:appendix-related-work}.

\providecommand{\mR}{\mathbb{R}}
\providecommand{\mZ}{\mathbb{Z}}
\providecommand{\mZp}{\mathbb{Z}_{+}}
\providecommand{\cX}{\mathcal{X}}
\providecommand{\cC}{\mathcal{C}}
\providecommand{\cF}{\mathcal{F}}
\providecommand{\pf}[1]{\mathcal{P}(#1)}
\providecommand{\pftrue}{\mathcal{P}^{\ast}}
\providecommand{\pdist}[2]{d_{p}\!\left(#1,#2\right)}
\providecommand{\numobj}{M}
\providecommand{\aspace}{[K]}
\providecommand{\armset}{[K]}
\providecommand{\vectle}{\preceq_{\cC}}
\providecommand{\vectl}{\prec_{\cC}}

\section{Formulation}
\label{sec:model-covar-shift}

In this section, we formalize the contextual multi-objective bandit problem under covariate shift, introduce the order induced by a cone, define Pareto sets (for both arms and policies), and set up the preference-based metric and regret.

\paragraph{Notation.}
For $n \in \mathbb{N}$, $[n] := \{1,2,\ldots,n\}$. 
We write $\mZ$ and $\mZp$ for the sets of non-negative and positive integers, respectively. 
The norms $\Vert \cdot \Vert_{1}$, $\Vert \cdot \Vert_{2}$, $\Vert \cdot \Vert_{\infty}$ denote the $\ell_{1}$-, $\ell_{2}$-, and $\ell_{\infty}$-norms. 
The set of arms is $[K]$ and the set of objectives is $[M]$.
For $z\in\mR^{\numobj}$, $z^{(m)}$ is its $m$-th component.
Let $e_m$ be the unit vector with a $1$ in position $m$.
We assume $\cX \subset \mR^{d}$ is a complete metric space.

\paragraph{Problem setup and cone order.}
Given a polyhedral cone $\cC$ (definitions/properties in Appendix~\ref{sec:appendix-cone-related}), the mean reward of arm $j\in[K]$ in context $X\in\cX$ is $\mu_j(X)\in\mR^{\numobj}$. 
Vectors in $\mR^{\numobj}$ are ordered w.r.t.\ $\cC$:
\begin{definition}[Partial order induced by $\cC$]
\label{defn:partial-order}
For $\mu,\mu'\in\mR^{\numobj}$, write $\mu \vectle \mu'$ if $\mu-\mu' \in \cC$ (weak order) and $\mu \vectl \mu'$ if $\mu-\mu' \in \mathrm{int}(\cC)$ (strict order).
\end{definition}

\begin{definition}[Order over arms]
\label{defn:arm-order}
Fix $X\in\cX$ and arms $i,j\in[K]$:
\begin{enumerate}
\item $i$ \emph{weakly dominates} $j$ iff $\mu_j(X)\vectle \mu_i(X)$.
\item $i$ \emph{dominates} $j$ iff $\mu_j(X)\vectle \mu_i(X)$ and $\mu_j(X)\neq \mu_i(X)$.
\item $i$ \emph{strongly dominates} $j$ iff $\mu_j(X)\vectl \mu_i(X)$.
\end{enumerate}
\end{definition}

\begin{definition}[Pareto arms and Pareto set]
An arm $i\in[K]$ is \emph{Pareto optimal} for context $X$ if it is not dominated by any other arm in $[K]$ with respect to $\cC$. 
The \emph{Pareto set} $\pf{X}$ is the set of mean reward vectors of all Pareto arms at $X$.
\end{definition}

\begin{figure}[htbp]
    \centering
    \begin{subfigure}[t]{0.48\textwidth}
        \centering
        \includegraphics[width=\linewidth]{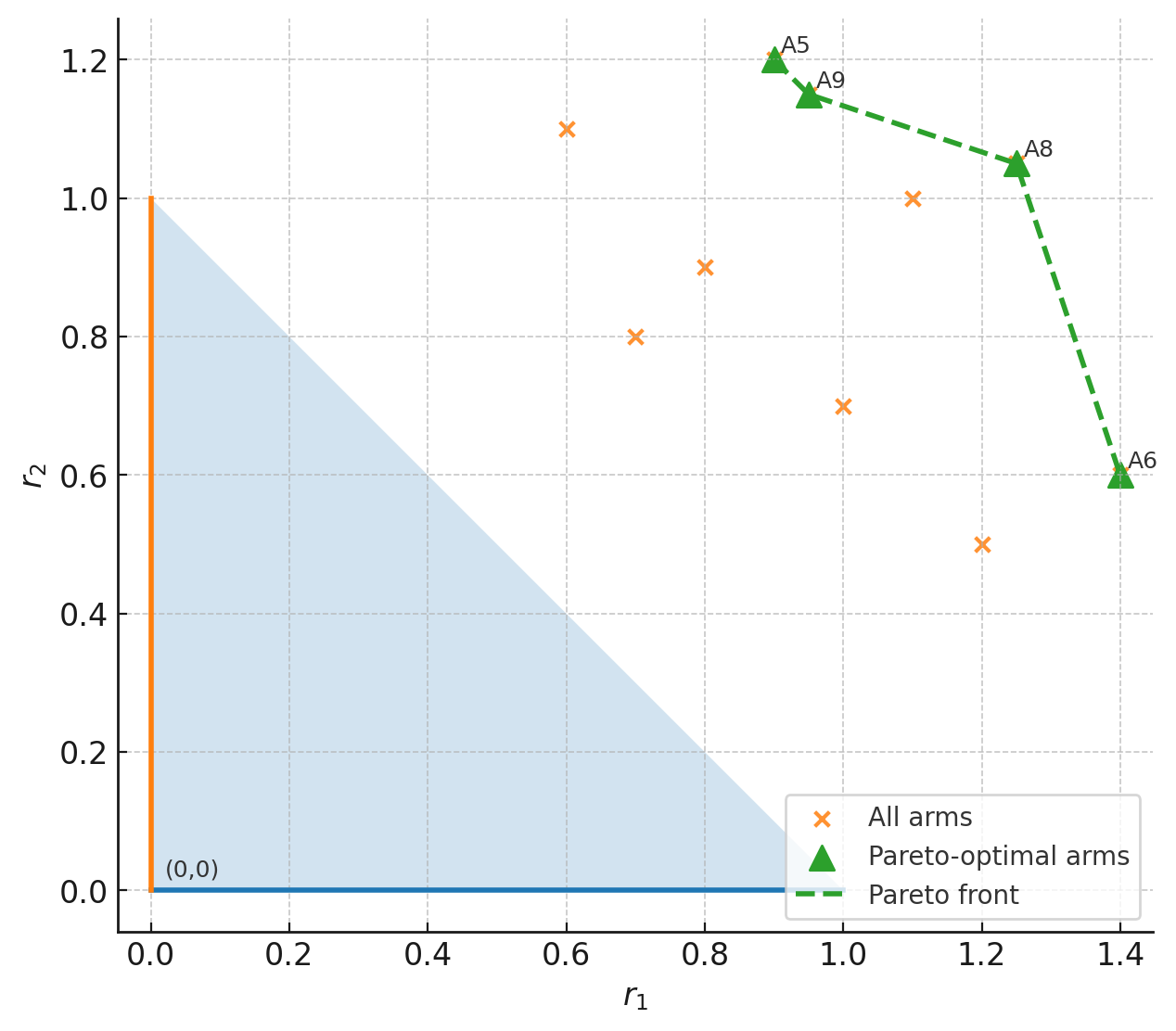}
        \caption{Positive orthant}
        \label{fig:cone_C1_formulation}
    \end{subfigure}
    \hfill
    \begin{subfigure}[t]{0.48\textwidth}
        \centering
        \includegraphics[width=\linewidth]{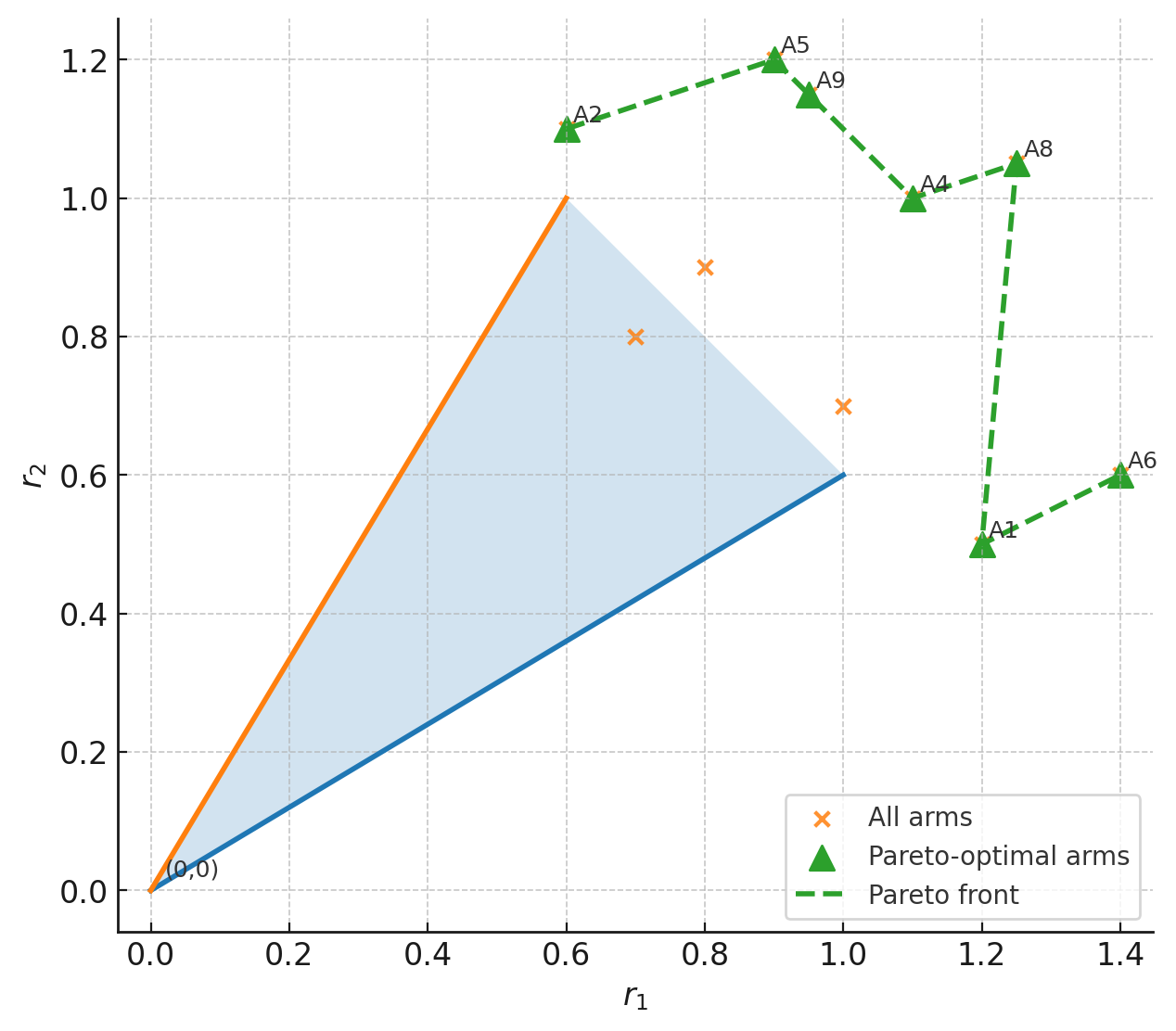}
        \caption{Narrower cone.}
        \label{fig:cone_C2_formulation}
    \end{subfigure}
    \caption{Pareto sets vary as the preference cone changes.}
    \label{fig:pareto_cone_grid}
\end{figure}

\paragraph{Covariate shift model.}
Let $T$ be the horizon and $t_p\le T$ be a fixed, unknown change-point. 
Contexts follow the \emph{source} distribution $P$ before $t_p$ and the \emph{target} distribution $Q$ after:
\[
X_t \sim \begin{cases}
P, & t \le t_p,\\
Q, & t > t_p.
\end{cases}
\]
At each $t$, after observing $X_t$, the learner selects $k_t\in[K]$ and observes a random reward vector $r_t\in\mR^{\numobj}$.

\begin{definition}[Dissimilarity measure {\cite{pathak2022new}}]
\label{defn:dissimilarity-metric}
For $h>0$,
\[
\rho_{h}(P,Q) \;=\; \int_{\cX} \frac{1}{P\!\left(B(x,h)\right)} \, dQ\!\left(B(x,h)\right),
\]
where $B(x,h)$ is the closed ball of radius $h$ centered at $x$.
\end{definition}
Larger $\rho_h(P,Q)$ indicates greater shift; the adversarial case corresponds to $\rho_h(P,Q)\to\infty$.

\paragraph{Reward model and noise.}
The $m$-th component of the reward is
\begin{equation}
\label{eqn:rew-vec}
r_t^{(m)} \;=\; \mu_{k_t}^{(m)}(X_t) + \eta_t^{(m)},\quad m\in[M],
\end{equation}
where $\{\cF_t\}$ is the natural filtration of the history $\{(X_s,k_s,r_s)\}_{s\le t}$, 
$\mathbb{E}[\eta_t^{(m)}\mid \cF_{t-1}]=0$, and $\eta_t^{(m)}$ is $\sigma$-sub-Gaussian:
$\mathbb{E}\!\left[\exp(\alpha \eta_t^{(m)})\mid \cF_{t-1}\right]\le \exp\!\left(\alpha^2\sigma^2/2\right)$ for all $\alpha>0$.

\paragraph{Policies and policy-induced Pareto sets.}
Let $\lambda([K])$ be the set of probability measures over $[K]$.
\begin{definition}[Family of policies]
\label{def:policy-family}
$\Pi := \big\{ \{\pi_t\}_{t=1}^{T} : \pi_t:\cX\to \lambda([K]),\; \pi_t \text{ is }\cF_t\text{-measurable} \big\}$.
\end{definition}
Given $\pi(X)\in\lambda([K])$, let $\mathrm{supp}(\pi(X)) := \{i\in[K]: \pi(X)(i)>0\}$.
\begin{definition}[Pareto set associated with a policy]
\label{def:policy-pareto}
For context $X$, the \emph{policy Pareto set} contains those $i\in \mathrm{supp}(\pi(X))$ that are not dominated (at $X$) by any other arm in $\mathrm{supp}(\pi(X))$.
We denote it by $\hat{\cP}^{\pi}(X)$ when based on estimates of means, and by $\cP^{\pi}(X)$ when based on the true means.
\end{definition}

\subsection{Preference-based gaps, metric, and regret}
We adopt a scale-independent notion of \emph{gap} and lift it to a metric on Pareto sets.

\begin{definition}[Scale–independent gap]
\label{defn:pareto-gap}
Fix $X\in\cX$ and an arm $k\in[K]$. 
Define
\[
\Delta\!\left(k,\pf{X}\right)
\;:=\;
\inf_{\varepsilon\in[1,\infty)^M}
\;\bigl\|\log \varepsilon\bigr\|_{\infty}
\quad\text{s.t.}\quad
\mu_k(X)\odot \varepsilon \not\prec_{\cC} \mu_{k'}(X)\quad
\forall\,k'\in \pf{X},
\]
where $\odot$ and $\log$ act component-wise, and $\not\prec_{\cC}$ means “not strictly dominated” under $\cC$.
\end{definition}

Intuitively, $\varepsilon$ is the smallest multiplicative factor (in any coordinate) needed so that $k$ is not strictly dominated
\noindent
\emph{Remarks.} 
(i) $\Delta(k,\pf{X})\ge 0$, with equality iff $k$ is Pareto optimal at $X$.
(ii) For $M=1$, $\Delta$ reduces to $|\log(\mu_k(X)/\mu_{k^\ast}(X))|$, a bona fide metric on $\mR_{>0}$. 
(iii) $\Delta$ is scale-independent: multiplying all objectives by the same $\alpha>0$ leaves $\Delta$ unchanged.

\paragraph{Hausdorff and its limitation here.}
The classical Hausdorff metric on sets with a base metric $d$ is
\[
d_H(A,B) \;=\; \max\left\{ \sup_{a \in A} \inf_{b \in B} d(a,b), \;
                           \sup_{b \in B} \inf_{a \in A} d(a,b) \right\}.
\]
However, even with a scale-independent base distance, Hausdorff convergence of Pareto sets can fail to reflect preference convergence.

\begin{exmp}[Hausdorff insufficiency]
\label{eg:insuff-hausdorff}
Let $\cX=[0,1]$, $K=3$, $M=2$ with
$\mu_{1}(x)=\begin{psmallmatrix}1\\0\end{psmallmatrix}$,
$\mu_{2}(x)=\begin{psmallmatrix}0\\1\end{psmallmatrix}$,
$\mu_{3}(x)=\begin{psmallmatrix}x^{2}\\x^{2}\end{psmallmatrix}$.
For $X_t=1/t$, $\pf{X_t}$ is the line segment joining 
$\begin{psmallmatrix}0\\1\end{psmallmatrix}$ to $\begin{psmallmatrix}t^{-2}\\t^{-2}\end{psmallmatrix}$ and 
$\begin{psmallmatrix}1\\0\end{psmallmatrix}$ to $\begin{psmallmatrix}t^{-2}\\t^{-2}\end{psmallmatrix}$.
As $t\to\infty$, $\pf{X_t}$ approaches the axes segment between $\begin{psmallmatrix}1\\0\end{psmallmatrix}$ and $\begin{psmallmatrix}0\\1\end{psmallmatrix}$, but the limit point $\begin{psmallmatrix}0\\0\end{psmallmatrix}$ is excluded. Thus, Hausdorff convergence does not imply convergence of induced preference relations.
\end{exmp}

\begin{definition}[Preference-based metric on Pareto sets]
\label{defn:pareto-metric}
For Pareto sets $\cP_1,\cP_2 \subset \mR^{\numobj}$, define
\[
\pdist{\cP_1}{\cP_2}
\;:=\;
\max\!\left\{
\sup_{k \in \cP_1} \Delta\!\big(k,\cP_2\big),\;
\sup_{k \in \cP_2} \Delta\!\big(k,\cP_1\big)
\right\}.
\]
\end{definition}
Intuitively, $d_p$ is the smallest worst-coordinate multiplicative adjustment (in log scale) needed so that each set does not strictly dominate the other. This parallels the Hausdorff distance but replaces the base metric with the scale-independent gap $\Delta_p$. It is shown to be a metric in Appendix~\ref{sec:prop-pareto-metric}.

\paragraph{Regret.}
We compare a policy to an oracle that knows $P,Q$, and $t_p$.
The \emph{preference-based regret} up to time $T$ is
\begin{equation}
\label{eqn:pareto_reg}
\cR(T)
\;=\;
\mathbb{E}\!\left[
\sum_{t=t_p+1}^{T} 
\pdist{\cP^{\pi}(X_t)}{\pftrue(X_t)}
\right],
\end{equation}
where the expectation is over the policy randomness and over $(P,Q)$-generated contexts. The regret quantifies the cumulative discrepancy between the policy-induced and oracle Pareto sets in terms of preference robustness. Note that although regret is summed over $t>t_p$, decisions at those times depend on information gathered under $P$. 

\begin{figure}[htbp]
    \centering
    \begin{subfigure}[t]{0.48\textwidth}
        \centering
        \includegraphics[width=\linewidth]{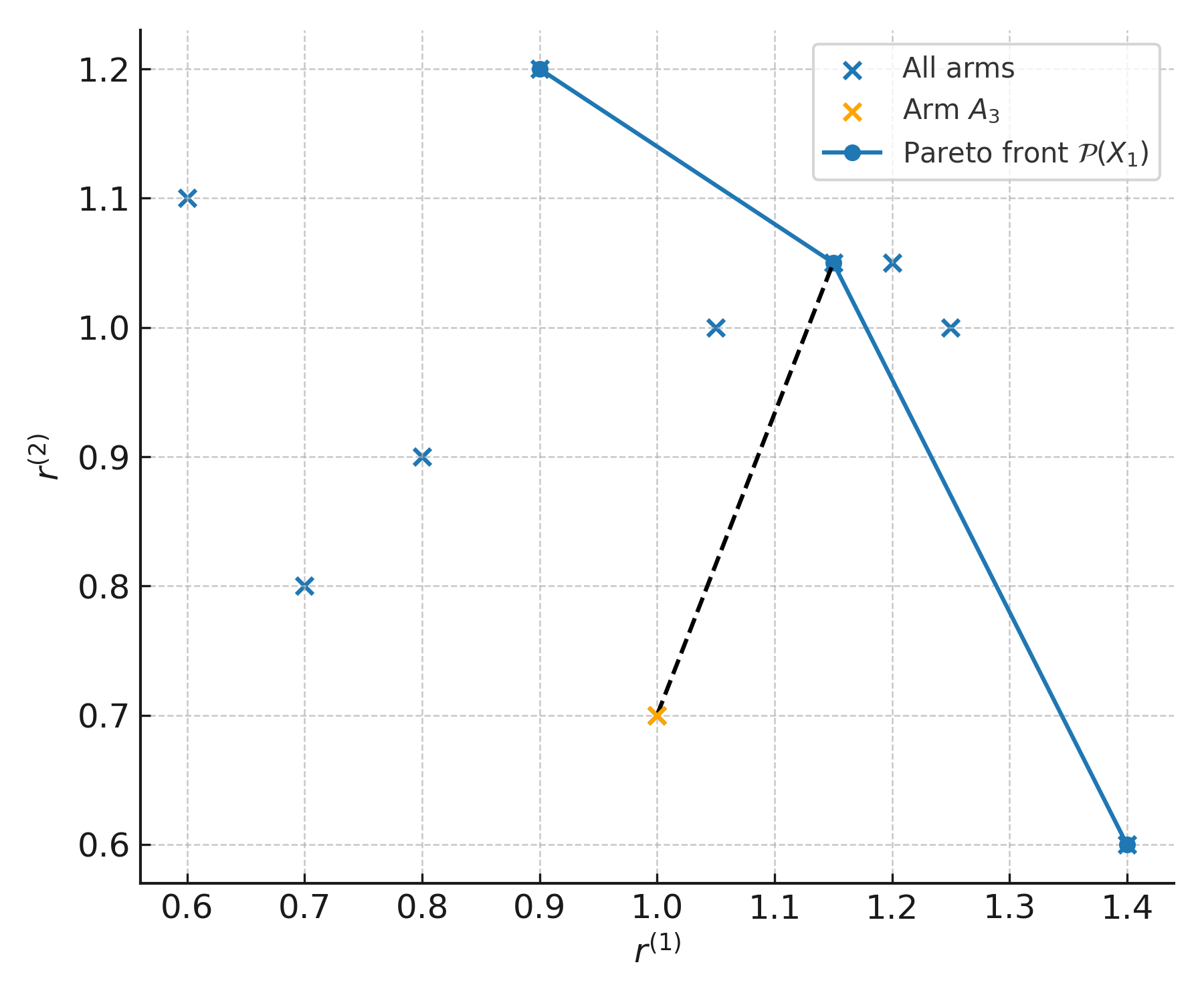}
        \caption{Gap}
        \label{fig:scale-gap}
    \end{subfigure}
    \hfill
    \begin{subfigure}[t]{0.48\textwidth}
        \centering
        \includegraphics[width=\linewidth]{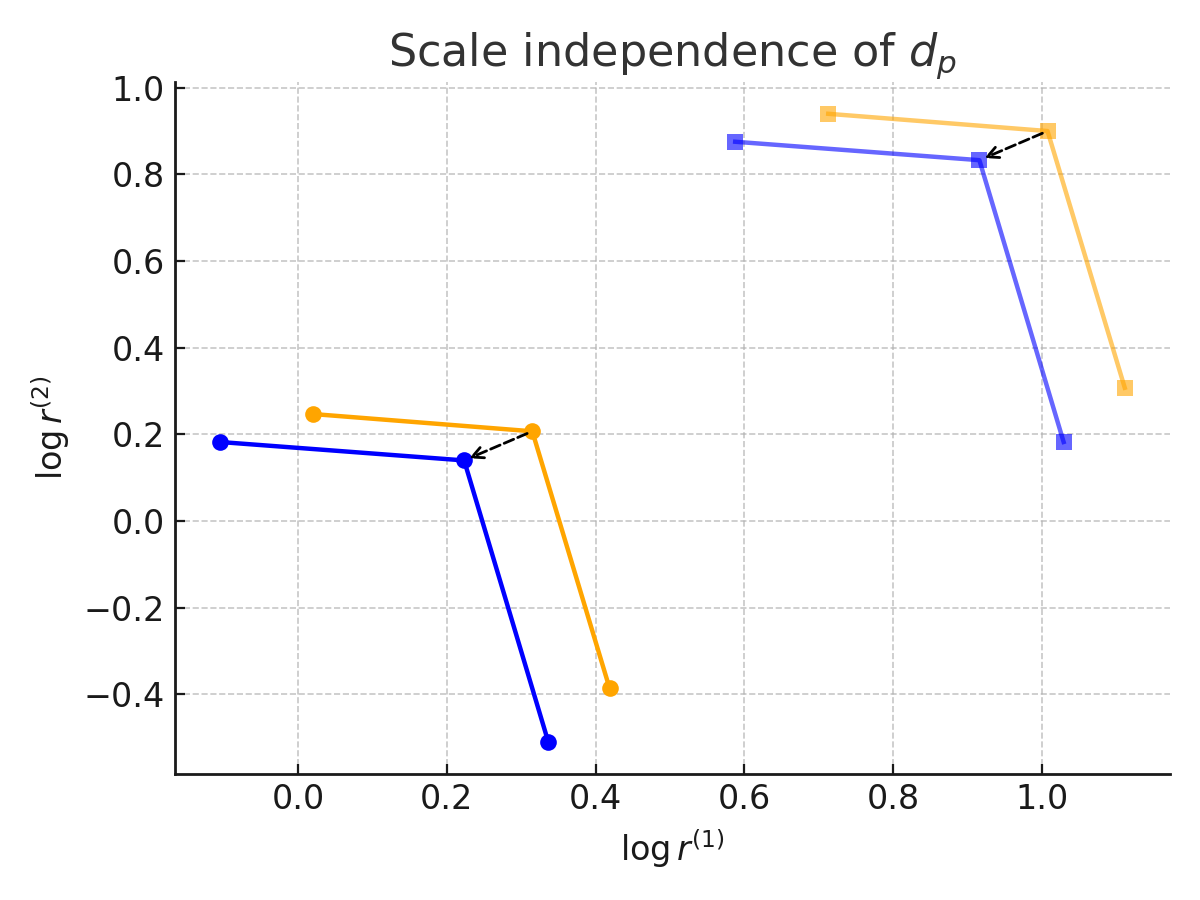}
        \caption{Preference metric}
        \label{fig:pref-metric}
    \end{subfigure}
    \caption{Schematic for Definition~\ref{defn:pareto-gap} and~\ref{defn:pareto-metric}.}
    \label{fig:pareto-metric}
\end{figure}

\subsection{Regularity assumptions and problem class}
\begin{assumption}[H\"older continuity in context]
\label{assmpt:holder_cont}
For all $k\in[K]$ and $X,X'\in\cX$,
\[
\max_{m\in[M]} \big|\mu_k^{(m)}(X)-\mu_k^{(m)}(X')\big|
\;\le\; C_\beta \,\|X-X'\|^{\beta}
\quad\text{for some } \beta\in(0,1],\ C_\beta>0.
\]
Let $\cH(\beta,C_\beta)$ denote this function class.
\end{assumption}

\begin{assumption}[Margin condition]
\label{assmpt:margin}
There exist $\epsilon_0,\alpha,C_\alpha>0$ such that for all $0<\epsilon\le \epsilon_0$ and all $X'\in\cX$,
\[
Q\!\left(\big\{X:\ \pdist{\pf{X}}{\pf{X'}} \le \epsilon\big\}\right)
\;\le\; C_\alpha \,\epsilon^{\alpha}.
\]
We denote by $\cM(\alpha,C_\alpha)$ the class of target distributions satisfying this condition.
\end{assumption}

\noindent
\emph{Discussion.} 
Assumption~\ref{assmpt:margin} controls the mass of contexts where Pareto sets are “$\epsilon$-close’’ under $d_p$, and is incomparable to Pareto zooming/packing conditions based on scalar sub-optimality gaps such as~\cite{turugay2018multi}.

\begin{definition}[Problem class]
\label{def:problem-class}
For horizon parameters $(t_p,T)\in \mZp^2$ with $t_p\le T$, define
\[
\Gamma(\alpha,C_\alpha,\beta,C_\beta,t_p,T)
\;:=\;
\big\{(P,Q,\mu):\ Q\in \cM(\alpha,C_\alpha),\ \mu\in \cH(\beta,C_\beta)\big\}.
\]
For brevity, we write $\Gamma$ when parameters are understood.
\end{definition}

\begin{definition}[Space of Pareto sets]
\label{defn:space-pareto-sets}
Let
\[
\paretospace
\;:=\;
\big\{\pf{X}:\ X\in\cX,\ \mu_k(\cdot)\in \cH(\beta,C_\beta)\ \text{for all }k\in[K]\big\}.
\]
\end{definition}

\section{Proposed Policy}

\begin{algorithm}
\caption{Preference-based contextual bandits under distribution shift}
\label{algo:cpf}
\begin{algorithmic}[1]
  \STATE{Input: Tree Partition $\treepart$ of context space $\cX$, Time horizon $\horizon$, arm set $\armset$}
  \WHILE{$1 \le t \le \horizon$}
  \label{algo-line:start-explore}
  \STATE{Observe the context $X_t$}
  \IF{$t < 8K\log\frac{KL}{\delta}$}
  \STATE{Play $k_t \leftarrow t \mod K +1$}
  \label{algo-line:uniform-arm-play}
  \STATE{Update estimates of mean-reward}
  \label{algo-line:end-explore}
\ELSE
   \STATE{Identify bin $(h_t,i_t)$ such that $X_t \in \bin{h_t}{i_t}, \bin{h_t}{i_t} \in \leaft$}
   \label{algo-line:bandwidth-selection}
\STATE{Initialise active arms: $\activearm \leftarrow \cap_{(h,i) \in \text{Pa}(h_t,i_t)} \cA_{(h,i)}$}
\STATE{$\pfestm:=\big\{k \in \activearm$: $u_{k,t}\binid \ \notdom \ u_{k',t}\binid, \ \forall \ k' \in \activearm$ \big\}}
\label{algo-line:pareto-front}
\STATE{Refine active arms: $\activearm\leftarrow \left\{k \in \activearm: \big\vert \hat{\mu}_{k,t} - \hat{\mu}_{k',t} \big\vert \vectle 2C_{t,h_t}, \ k' \in \pfestm\right\}$}
\label{algo-line:active-arm-refine}
\STATE{Play the arm $k_t$ uniformly at random from $\cA_{(h_t,i_t)}$}
\label{algo-line:play-random-arm}
\STATE{Update estimates for arms $k \in \cA_{(h_t,i_t)}$ using~\eqref{eqn:stat-uncert}}
\label{algo-line:update-estimates}
\IF{$\sqrt{\frac{8K\log(\frac{KL}{\delta})}{n_{k,t}}} < V^{\beta}_h$}
\label{algo-line:bin-expansion}
\STATE{Add to leaf set: $\mathcal{L}_{t+1} \leftarrow \leaft \cup \text{child}(\bin{h_t}{i_t})$}
\STATE{Remove from leaf set: $\mathcal{L}_{t+1} \leftarrow \leaft \setminus \bin{h_t}{i_t}$}
\ELSE
\STATE{$\mathcal{L}_{t+1} \leftarrow \leaft$}
\ENDIF
\ENDIF
\ENDWHILE
\end{algorithmic}
\end{algorithm}

Recall that the performance of our algorithm is measured with respect to an oracle that knows the change point $t_p$, the source and target distributions. Given this information, the oracle knows the arms to be played in order to minimize regret irrespective of the distribution from which the adversary draws the contexts. However, the decision maker lacks this information and therefore, needs to adapt to this shift. We propose a tree-based discretization policy that automatically adapts to this shift in context distribution. The policy discretizes the context space using a dyadic tree. We assume that we have a (tree) structured partitioning $\tree$ for the context space $\cspace$ defined as follows.
\begin{definition}[Tree Partition]
\label{defn:tree-partition}
A (tree) structured partitioning $\tree$ of $\cspace$ is a doubly indexed collection of
bins (subsets) $\{\bin{h}{i} \subset \cspace: h \in \mN, i = 1, \ldots, \Psi^h\}$, for some  $\Psi \in \mZ_{+}$, that satisfies the following conditions: 
  \begin{enumerate}
  \item The root $\tree_0:= \bin{0}{1} = \cspace$
  \item \label{def:tree:children} For $\numchild \ge 1$, we have that
    $\tree_h := \{\bin{h}{i} \subset \cX: i= 1, \ldots, \Psi^h\}$. Then:
    \begin{enumerate}
    \item For fixed $h \ge 0$, each $\tree_h$ is a partition of $\cX$, i.e., $\cup_{i=1}^{\Psi^h} \bin{h}{i} =
      \cX$, and $\bin{h}{i} \cap \bin{h}{j} = \emptyset$ for $i
      \neq j$.
    \item For each $h \ge 0$, $\cT_{h}$ is a tree, i.e. $\bin{h}{i} =
      \cup_{j = (i-1)\Psi+1}^{i\Psi} \bin{h+1}{j}$. 
    \end{enumerate}
    We will call sets $\left\{\bin{h+1}{(i-1)\Psi + j)}: j = 1, \ldots,
      \numchild \right\}$ the \emph{children} of the set $\bin{h}{i}$, and define
    the \emph{parent} $\text{Pa}(h+1,(i-1)\Psi+j) = (h,i)$ for all $j = 1, \ldots, \numchild$. The ancestors of the bin are defined as: 
    \[
    \anc{\bin{h}{i}} = \left\{\bin{h}{i}': \bin{h}{i}' \in (\text{Pa}({\bin{h}{i}}))^{k}, \ k \ge 1\right\} 
    \] 
  \item \label{def:tree:width} Let $0 < \binrad < 1$ denote the width, i.e., length of side of hyperrectangle of bin $\bin{h}{i}$ at level $h$.
  \item \label{def:tree:repElement} Let $X_{(h,i)}$ denote the geometric center of $\bin{h}{i}$. For each bin $\bin{h}{i}$ we maintain a set of active arms $\cA_{\binid}$ such that $\cA_{\binid} \subseteq \cA_{\text{Pa}(\binid)} \subseteq \armset$.
  \end{enumerate}
\end{definition}

A bin in the tree $\bin{h}{i}$ is specified by a pair of indices $(h,i)$ with the interpretation that this bin corresponds to the set $\bin{h}{i}$ and that all problem and algorithmic attributes are evaluated at the geometric center of the bin unless otherwise specified. Any policy can reach only a finite depth of the tree when run for a finite-time period. Let $\leaf_{t}$ denote the set of leaves defined as bins that have been visited in the past by the policy but whose children have not been explored until time $t$. 

\subsection{Policy}
At any time $t$, in addition to determining the arm to be played, any policy $\pi$ faces the exploration-exploitation dilemma of either splitting a leaf bin into children or exploring information about mean rewards associated with this bin. These two ingredients are inter-related through selection of a bin upon observing a context and using statistical estimates for arm selection and bin exploration. We detail them separately below.

\textbf{Bin selection and exploration:} Starting from the current set of leave bins of $\tree$, upon observing a context $X_t$, the policy picks a bin $(h_t,i_t) \in \leaft$. The policy starts with the root bin $\bin{0}{1}$ and leaf set $\cL_{0} := \bin{0}{1}$. The policy begins by playing each arm in a round-robin fashion to generate initial estimates of the mean reward for the arm set in the root bin $\cB_{(0,1)}$. It then grows by partitioning one of the bins in the current leaf set into its children bins. The decision of when to split the bin depends on the uncertainty associated with the estimates of the mean rewards. For this purpose, for bin indexed $(h,i)$, let $n_{k,t}\binid$ denote the number of times a context falls into the bin $(h,i)$ and arm $k$ is played.
\begin{equation}
  \label{eq:n-def}
  n_{k,t}(h,i) := \sum_{s = 1}^t \ind{X_s \in \bin{h}{i}, k_s = k} 
\end{equation}
and $\hat{\mu}_{k,t}\binid$ denote the empirical estimate of the mean reward for that bin based on the regressogram (Line~\ref{algo-line:update-estimates} in Algorithm~\ref{algo:cpf}):
\begin{eqnarray}
\label{eq:muhat-hi-def}
\hat{\mu}_{k,t}(h,i) &:=& \frac{\sum_{s = 1}^{t-1} r_s\ind{x_s \in \bin{h}{i}, k_s = k}}{\sum_{s = 1}^t \ind{x_s \in \bin{h}{i}, k_s = k}},  
\end{eqnarray}
Using Assumption~\ref{assmpt:holder_cont}, the error in estimates of mean reward of any context in a bin can be decomposed in terms of the expected reward of that context and deviation from this value which is representative of uncertainty in the estimate (see Appendix~\ref{appendix:thm1}). For $\delta \in (0,1)$ is to be specified later, denote the upper confidence bound associated with any bin at level $h$ when arm $k$ is played $n_{k,t}$ number of times (see Appendix~\ref{appendix:thm1}), is given by: 
\begin{eqnarray} 
\label{eqn:stat-uncert}
\uncertrad := c_{1}V_{h}^{\beta} + \sqrt{\frac{\log\left(\frac{KM}{\delta}\right)}{n_{k,t}}}
\end{eqnarray}
The optimistic upper bound for reward function in bin $\bin{h}{i}$ is given by:
\begin{align}
\label{eq:u-def}
u_{k,t}(h,i) :=\hat{\mu}_{k,t}+\uncertrad
\end{align}

As the tree level increases, the center of the bin provides finer approximation of the mean reward for any context in that bin. The minimum uncertainty in the estimates of the mean reward in a particular bin is at least the width of the bin. In order to allow for a continuous improvement in estimates of the mean rewards for active arms in that bin, a bin is partitioned when the uncertainty due to sampling is lower than bin width. When the stochastic error in the estimate of the reward function $\hat{\mu}_{k,t}\currbin$ of bin $(h_t,i_t)$ is less than the width of the bin $\bin{h}{i}$, the bin is split into children bins. When a bin $\cB\currbin$ is partitioned, its children are initialized into the set of active leaves (and endowed with mean reward estimates and set of active arms from their parents). The bin itself is removed from the set of active leaves.

\textbf{Arm Selection:}
Arms at time $t$ are selected from a set of active arms $\cA_{(h_t,i_t)}$ for bin $\currbin$. The active arms contain the set of Pareto optimal arms for the true mean reward for bin $\currbin$ with high probability (see Appendix~\ref{appendix:thm1}). When a bin is initialized, each bin is endowed with a set of active arms from its parent bin (the active set at the root is the entire set of arms). The set of active arms is based on estimates of the mean reward~\eqref{eq:u-def}. They are further refined as more samples are observed and deeper levels of the tree are explored. This creates a finer approximation of the mean reward and shrinks the radius of the confidence ball around estimates of the mean rewards. The arms are eliminated once they are determined not to belong to the Pareto front associated with the true mean rewards in this bin. An arm is eliminated if the relative gaps between the estimated mean rewards for different arms is large enough. From the set of active arms, the policy constructs a Pareto front based on estimated mean-rewards. An arm is then played with uniform probability from this estimated Pareto front.

\textbf{Other insights:}
Our policy is an adaptation of elimination-based adaptive-discretization algorithm for regret minimization. It self-tunes to several unknown parameters such as the change-point $t_p$ and the margin parameter $\alpha$, the dissimilarity metric $\rho$ and the context dimension $d$. First, past work has focused on tree-based discretization policies wherein the estimator converges at the optimal regression rate, $t^{\frac{-1}{2+d}}$. Such a strategy however, does not work under distribution shift. Second, the optimal choice of tree-level is further complicated under distributional shift due dissimilarity between source $P$ and target $Q$, which would scale as $\rho(P,Q)^{\alpha}$. Adaptation to the unknown margin parameter $\alpha$ comes through such decisions over the set of active arms. Namely, if the margin is much larger than the width of a bin, then all suboptimal arms are discarded quickly so we suffer no regret for playing arms in that bin. On the other hand, for low-margin regions, we can bound the regret due to playing active arms. Furthermore, Assumption~\ref{assmpt:margin} ensures that the probability of a covariate landing in bins with a small margin is low. Another technical detail is that we don’t constrain the covariate distribution to follow a strong density assumption or near-uniformity. As a result, the regret due to variance in estimation can be large due to the random choice of $X_t$ since the different bins at the same tree-level can have different density under $Q$. A careful peeling argument
integrating these subtleties is the main technical challenge for our regret analysis. 

\textbf{Numerical Experiments:}
We evaluate the performance of Algorithm~\ref{algo:cpf} by benchmarking its performance under covariate shift on synthetic  instances. We show that the behavior of regret with problem parameters is as reported in our theoretical results. These experimental studies are reporeted in Appendix~\ref{appendix:numerical-experiments} due to space constraints.

\section{Regret Analysis}
\label{sec:pareto_continuum}

We now bound the regret of Algorithm~\ref{algo:cpf} under various distribution shift scenarios.  
Our first result considers the case of a \emph{single} change point.

\begin{theorem}[Single Shift]
\label{thm:cs-regret}
Under Assumptions~\ref{assmpt:holder_cont}--\ref{assmpt:margin}, when Algorithm~\ref{algo:cpf} is run on an instance from $\Gamma$, with probability at least $1-\delta$ the regret satisfies
{\scriptsize
\begin{eqnarray*}
\mathcal{R}(T) &\le& \mathcal{O} \Bigg(
\left(\frac{K \log\left( \frac{KM}{\delta} \right)}{\max\{t_p,\, T-t_p\}} \right)^{\frac{\alpha + 1}{\beta}} \\
&&\quad + \left[ K \log\left( \frac{KM}{\delta} \right) \min\left\{ 
\frac{ \rho(\source, \target) }{t_p}, 
\frac{ \rho(\target, \target) }{T-t_p} 
\right\} \right]^{ \frac{\alpha + 1}{\alpha} \cdot \frac{\beta + 1}{\beta} }
\Bigg).
\end{eqnarray*}
}
\end{theorem}

\noindent\textbf{Discussion.}  
The first term corresponds to the exploration cost within each stationary phase (before/after the change point), scaling with $\beta$ via the smoothness of the reward function.  
The second term captures the adaptation cost due to distribution shift, proportional to the dissimilarity $\rho(\cdot,\cdot)$.  

\begin{remark}[Dependence on problem parameters]
\label{remark:problem-parameter}
In the absence of covariate shift ($t_p=0$), Theorem~\ref{thm:cs-regret} reduces to
{\scriptsize
\begin{eqnarray*}
\mathcal{R}(T) &\le& \mathcal{O} \Bigg(\left( \frac{K \log\left( \frac{KM}{\delta} \right)}{T} \right)^{\frac{\alpha + 1}{\beta}} + \left[ K \log\left( \frac{KM}{\delta} \right)  \frac{ \rho(\target, \target) }{T} \right]^{ \frac{\alpha + 1}{\alpha} \cdot \frac{\beta + 1}{\beta} }\Bigg).
\end{eqnarray*}
}
This recovers the standard $T^{-(\alpha+1)/\beta}$ scaling of contextual bandits.  
The bound improves as $\beta \to \infty$ and worsens as $\alpha \to 0$.
\end{remark}

\subsection{Specialized source--target families}

We now specialize to a tractable and practically relevant family of source--target distributions, namely those satisfying a tree-based dissimilarity decay.

\begin{assumption}[Tree-discretized family]
\label{assmpt:family-dissimilarity}
The source and target distributions $(P,Q)$ satisfy
\[
\sup_{0 < h \le 1} h^{-\gamma} \rho_{h}(P,Q) \le C_{\gamma}
\]
for some $\gamma > 0$.  
We denote the family of such pairs by $\mathcal{D}(\gamma,C_{\gamma})$.
\end{assumption}

\begin{exmp}[Examples of $\mathcal{D}(\gamma,C_{\gamma})$]
\label{eg:tree-distribution}
If $P(x) = (1+\gamma) x^{-\gamma}$ (power law) and $Q$ is uniform on $[0,1]$, then $(P,Q) \in \mathcal{D}(\gamma,1+\gamma)$.  
Another example is an exponential distribution paired with a power-law distribution, which arises in modeling gene expression evolution.
\end{exmp}

\begin{theorem}[Special Families of Source--Target Pairs]
\label{thm:special-family}
Under Assumptions~\ref{assmpt:holder_cont}--\ref{assmpt:margin} and~\ref{assmpt:family-dissimilarity}, when Algorithm~\ref{algo:cpf} is run on $(P,Q) \in \mathcal{D}(\gamma,C_\gamma)$, with probability at least $1-\delta$ the regret satisfies
{\scriptsize
\begin{eqnarray*}
\mathcal{R}(T) &\le& \mathcal{O} \Bigg( 
\left[ K \log\left( \frac{KM}{\delta} \right) \min\left\{ \frac{1}{t_p}, \frac{1}{T-t_p} \right\} \right]^{\frac{\alpha+1}{\alpha} \cdot \frac{\gamma(\beta+1)}{\beta}}
\left( \frac{K \log\left( \frac{KM}{\delta} \right)}{T-t_p} \right)^{\frac{1}{\alpha}} \\
&&\quad + \left[ K \log\left( \frac{KM}{\delta} \right) \min\left\{ \frac{1}{t_p}, \frac{1}{T-t_p} \right\} \right]^{\frac{\alpha+1}{\beta}}
\Bigg).
\end{eqnarray*}
}
\end{theorem}

\noindent\textbf{Discussion.}  
Compared to Theorem~\ref{thm:cs-regret}, the adaptation term now carries the factor $\gamma/\beta$, reflecting how quickly dissimilarity decays in the tree metric relative to reward smoothness.  

\subsection{Multiple shifts}

In many applications, the context distribution shifts multiple times:
\[
P_1 \to P_2 \to \dots \to P_{n} \to Q,
\]
at (unknown) change points $t_1, t_2, \dots, t_n$.  
Our algorithm handles this scenario without modification.

\begin{theorem}[Multiple Shifts]
\label{thm:multiple-shift}
Under Assumptions~\ref{assmpt:holder_cont}--\ref{assmpt:margin}, suppose the total number of source samples is $t_p = \sum_j t_j$, where $t_j$ is the duration under source $P_j$.  
Let $\tilde{P}$ be the mixture $\tilde{P} = \sum_{j} \frac{t_j}{t_p} P_j$.  
Then with probability at least $1-\delta$,
{\scriptsize
\begin{eqnarray*}
\mathcal{R}(T) &\le& \mathcal{O}\left(\frac{K\log\frac{KM}{\delta}}{T-t_p}\right)^{\frac{1}{\alpha}}
\left(\frac{K \log\left(\frac{KM}{\delta}\right)}{\max\{t_p,T-t_p\}}\right)^{\frac{\alpha+1}{\beta}} \\
&&\quad + \left[ K\log\left(\frac{KM}{\delta}\right)\min\left\{\frac{\rho(\tilde{P},\target)}{t_p},\frac{\rho(\target,\target)}{T-t_p}\right\}\right]^{\frac{\alpha+1}{\alpha} \cdot \frac{\beta+1}{\beta}}.
\end{eqnarray*}
}
\end{theorem}

\noindent\textbf{Discussion.}  
The bound depends on $\rho(\tilde{P},Q)$, the dissimilarity between the target and the mixture of sources.  
In geometric terms, the intermediate distributions from $P_1,\dots,P_n$ “average out” into a single effective distribution before aligning with $Q$.  
When $n=1$, $\tilde{P}=P$ and this reduces to Theorem~\ref{thm:cs-regret}.

\begin{proof}[Proof sketch]
We use that $\rho(\tilde{P},\target)$ is the weighted average of the dissimilarities between each $P_j$ and $Q$.  
Convexity arguments then reduce the bound to the form of Theorem~\ref{thm:cs-regret}.
\end{proof}

\section{Conclusion}
We considered the contextual bandit problem under distribution shift and vectorial reward functions. We proposed an adaptive discretization and OFU based learning policy. In order to quantify the performance of such a policy, we introduced a metric over the space of Pareto fronts and analysed regret under this metric. An interesting future work would be to consider this problem for the kernelized contextual bandit framework. 

\bibliographystyle{apalike}
\bibliography{shared/ref}

\newpage 
\appendix
\part{Appendix}
\section{Preliminaries and definitions}
\label{sec:appendix-cone-related}
\begin{definition}[Cone]
A set $C\subseteq \mathbb{R}^n$ is a (convex) \emph{cone} if for every $x,y\in C$ and
$\alpha,\beta\ge 0$ we have $\alpha x+\beta y\in C$. Equivalently, $x\in C$ and
$\lambda\ge 0$ imply $\lambda x\in C$.
\end{definition}

\begin{definition}[Polyhedral cone]\label{def:polyhedral-cone}
A set $C\subseteq \mathbb{R}^n$ is a \emph{polyhedral cone} if it can be written as the
intersection of finitely many closed halfspaces whose bounding hyperplanes pass through
the origin; i.e., there exists $A\in\mathbb{R}^{m\times n}$ such that
\[
C \;=\; \{\, x\in\mathbb{R}^n : A x \le 0 \,\}.
\]
\end{definition}

\begin{definition}[Hausdroff Metric]
\label{defn:hausdroff-metric}
Let $(X,d)$ be a metric space and let $A,B \subset X$ be non-empty. 
The \emph{Hausdorff distance} between $A$ and $B$ is defined as
\[
d_H(A,B) \;=\; \max\left\{ \sup_{a \in A} \inf_{b \in B} d(a,b), \;
                           \sup_{b \in B} \inf_{a \in A} d(a,b) \right\}.
\]
\end{definition}

\begin{remark}[Finite generation (Minkowski–Weyl for cones)]
Equivalently, $C$ is a polyhedral cone iff there exist vectors
$v_1,\dots,v_k\in\mathbb{R}^n$ such that
\[
C \;=\; \operatorname{cone}\{v_1,\dots,v_k\}
\;:=\;
\Bigl\{\, \sum_{i=1}^k \lambda_i v_i \;:\; \lambda_i \ge 0 \Bigr\}.
\]
\end{remark}

\section{Figure~\ref{fig:pareto_cone_grid}}
\newcommand{\PF}{\mathrm{PF}}
\paragraph{Arms (K=10, M=2).}
\[
\begin{array}{c|cc}
\text{Arm} & r_1 & r_2 \\
\hline
A0 & 0.80 & 0.90 \\
A1 & 1.20 & 0.50 \\
A2 & 0.60 & 1.10 \\
A3 & 1.00 & 0.70 \\
A4 & 1.10 & 1.00 \\
A5 & 0.90 & 1.20 \\
A6 & 1.40 & 0.60 \\
A7 & 0.70 & 0.80 \\
A8 & 1.25 & 1.05 \\
A9 & 0.95 & 1.15 \\
\end{array}
\]

\paragraph{Cones and generating rays.}
\[
\begin{aligned}
C_1 &= \mathbb{R}^2_{+}
    = \cone\{\,w_1=(1,0),\; w_2=(0,1)\,\},\\[3pt]
C_3 &= \cone\{\,w_1=(1,0.6),\; w_2=(0.6,1)\,\}
    = \{\,W\alpha:\alpha\ge 0\,\},\quad
W=\begin{bmatrix}1 & 0.6\\[2pt] 0.6 & 1\end{bmatrix},\\
W^{-1} &= \frac{1}{1-0.6^2}\!\begin{bmatrix} 1 & -0.6\\ -0.6 & 1\end{bmatrix}
      = \begin{bmatrix} 1.5625 & -0.9375\\[2pt] -0.9375 & 1.5625 \end{bmatrix}.
\end{aligned}
\]
Dominance test:
\[
\text{Under }C_1:\; y \succeq x \iff y_1\!\ge x_1,\; y_2\!\ge x_2,\; \text{and } y\ne x.
\\
\text{Under }C_3:\; y \succeq x \iff W^{-1}(y-x)\ge 0 \text{ (componentwise), } y\ne x.
\]

\paragraph{Pareto front under \(C_1\) (orthant, maximization).}
\[
\PF_{C_1} = \{A5,\;A6,\;A8,\;A9\}.
\]
\emph{Dominance:}
\[
\begin{aligned}
&\text{A0 }(0.80,0.90)\;\text{is dominated by A9 }(0.95,1.15)\ \text{and A5 }(0.90,1.20)\ (\text{both coords }\uparrow).\\
&\text{A1 }(1.20,0.50)\;\text{is dominated by A6 }(1.40,0.60)\ \text{and A8 }(1.25,1.05).\\
&\text{A2 }(0.60,1.10)\;\text{is dominated by A5 }(0.90,1.20)\ \text{and A9 }(0.95,1.15).\\
&\text{A3 }(1.00,0.70)\;\text{is dominated by A4 }(1.10,1.00)\ \text{and A8 }(1.25,1.05).\\
&\text{A4 }(1.10,1.00)\;\text{is dominated by A8 }(1.25,1.05).\\
&\text{A7 }(0.70,0.80)\;\text{is dominated by many (e.g., A0, A4, A5, A8, A9).}\\
&\text{A5, A6, A8, A9 are each not dominated by any other arm under }C_1.
\end{aligned}
\]



\paragraph{``Pareto set under $C_1$''.}
\begin{itemize}
  \item \textbf{Cone:} $K_1=\{\alpha w_1+\beta w_2:\alpha,\beta\ge0\}$ with 
        $w_1=\begin{bmatrix}1\\0\end{bmatrix}$, 
        $w_2=\begin{bmatrix}0\\1\end{bmatrix}$.
  \item \textbf{Order:} $x \preceq_{K_1} y \iff (y-x)\in K_1$ (maximization).
  \item \textbf{Arms:} $\mathcal{A}=\{A_0=(0.8,0.9),\,A_1=(1.2,0.5),\,A_2=(0.6,1.1),\,A_3=(1.0,0.7),\,A_4=(1.1,1.0),\,A_5=(0.9,1.2),\,A_6=(1.4,0.6),\,A_7=(0.7,0.8),\,A_8=(1.25,1.05),\,A_9=(0.95,1.15)\}$.
  \item \textbf{Pareto set under $C_1$:} $\{A_5,\,A_6,\,A_8,\,A_9\}$.
  \item \textbf{Plotting:} shaded wedge = cone $K$; rays labeled $w_1,w_2$; triangles = Pareto; crosses = dominated.
\end{itemize}

\paragraph{``Pareto set under $C_2$''.}
\begin{itemize}
  \item \textbf{Cone:} $K_2=\{\alpha w_1+\beta w_2:\alpha,\beta\ge0\}$ with 
        $w_1=\begin{bmatrix}1\\0.6\end{bmatrix}$, 
        $w_2=\begin{bmatrix}0.6\\1\end{bmatrix}$.
  \item \textbf{Order:} $x \preceq_{K_2} y \iff (y-x)\in K_2$ (maximization).
  \item \textbf{Pareto set under $C_2$:} $\{A_1,\,A_2,\,A_4,\,A_5,\,A_6,\,A_8,\,A_9\}$.
  \item \textbf{Plotting:} same conventions as above; title intentionally shows $C_2$.
\end{itemize}

\begin{table}
\centering
\caption{Scale-independent gap $\Delta$ to the Pareto front under cones $C_1$ (orthant) and $C_3$ (narrow cone).}
\label{tab:gap_C1_vs_C3}
\begin{tabular}{lrrllrrrrr}
\toprule
Arm &   r1 &   r2 &  Pareto under C1? &  Pareto under C3? &  Delta\_C1 (log-gap) &  Delta\_C3 (log-gap)  \\
\midrule
 A0 & 0.80 & 0.90 &             False &             False &              0.1719 &              0.0000         \\
 A1 & 1.20 & 0.50 &             False &              True &              0.1542 &              0.0000 \\
 A2 & 0.60 & 1.10 &             False &              True &              0.0870 &              0.0000  \\
 A3 & 1.00 & 0.70 &             False &             False &              0.2231 &              0.0392 \\
 A4 & 1.10 & 1.00 &             False &              True &              0.0488 &              0.0000  \\
 A5 & 0.90 & 1.20 &              True &              True &              0.0000 &              0.0000  \\
 A6 & 1.40 & 0.60 &              True &              True &              0.0000 &              0.0000 \\
 A7 & 0.70 & 0.80 &             False &             False &              0.3054 &              0.0556 \\
 A8 & 1.25 & 1.05 &              True &              True &              0.0000 &              0.0000 \\
 A9 & 0.95 & 1.15 &              True &              True &              0.0000 &              0.0000  \\
\bottomrule
\end{tabular}
\end{table}

\section{Notation}
\small{\begin{longtable}{|l|l|l|}
\hline
Notation & Description & Comments \\
\hline
$[K],k_t$ & Set of arms, arm played at time $t$ & \\
\hline
$\cX_{t},\cX$ & Context at time $t$ and set of contexts & \\
\hline
$\mu_{k},\mu_{k'}$ & Mean reward for arms $k$ and $k'$ & \\
\hline
$T,t_{p}$ & Decision horizon and change time & \\
\hline
$\pftrue, \cZ$ & Pareto front associated with context $\cX$ and space of Pareto fronts on $\armset$ & \\
\hline
$\source,\target$ & Source and target context distributions &  \\
\hline
$\rho(\source,\target)$ & Dissimilarity metric &  Definition~\ref{defn:dissimilarity-metric} \\
\hline
$r_t,\eta_t$ & Reward vector and observation noise at time $t$ & See~\eqref{eqn:rew-vec} \\
\hline
$\pdist{\pfestm}{\pftrue}$& Distance between true and estimated Pareto fronts & Definition~\ref{defn:pareto-metric}\\
\hline
$\cR(T)$ & Regret at time $t$ & See~\eqref{eqn:pareto_reg} \\
\hline
$\beta,C_{\beta}$& H\"older continuity constants & Assumption~\ref{assmpt:holder_cont}\\
\hline
$\alpha,C_{\alpha}$ & Margin parameter and constant & Assumption~\ref{assmpt:margin}  \\
\hline
$\cT.\Psi$ & Tree partition of $\cX$ and number of childer per bin & Definition~\ref{defn:tree-partition}\\
\hline
$\leaft$ & Set of leaves &\\
\hline
$(h_t,i_t),\activearm$ & Bin selected at time $t$, set of active arms & \\
\hline
$\bin{h}{i}$ & Bin at $(h,i)$& \\
\hline
$\numkt$ & Number of times arm $k$ is played until time $t$ &  \\
\hline
$\muestm{\ell}{k}{\binid}$ & Mean reward estimate for objective $\ell$, arm $k$ at bin $\binid$ & \\
\hline 
$\Delta(x,k)$& Distance of mean reward of arm $k$ from Pareto front $\pftrue$ & Definition~\ref{defn:pareto-gap} \\ 
\hline 
$V_h$ & Bias in cell at depth $h$ & \\ \hline
$\cG$& Event that all arm means concentrate & Definition~\ref{defn:concentration-arm-mean} and Lemma~\ref{lem:arm-concentration} \\ \hline
$\cE$ & Event that covariate counts in the bins are sufficiently large & Lemma~\ref{lem:arm-covariate-count} \\ 
\hline
\caption{Notations}
\label{tab:Notation}
\end{longtable}
}

\section{Related Work}
\label{sec:appendix-related-work}
\begin{enumerate}
\item \textbf{Contextual and Continuum-armed Bandits:} In this paper, we consider the multi-armed problem where the expected reward is a non-parametric function of the observed context and action. One of the first such models considered a finite-armed bandit problem and was proposed by \citep{yang2002randomized}. This problem was subsequently studied by~\citep{rigollet2010nonparametric} and~\citep{perchet2013multi}. In particular,~\citep{rigollet2010nonparametric} analyzes an upper-confidence bound-based regressogram policy utilizing a static discretization of the context space. \citep{perchet2013multi} extended their analysis by proposing a policy when the context space can be adaptively discretized. Smoothness assumptions of these prior works was recently relaxed by~\citep{gur2019smoothness} and~\citep{hu2020smooth} who propose policy adaptive to the smoothness of the underlying mean-reward function.  In contrast to finite-armed bandit model, we consider the case when the set of arms forms a continuum. The continuum-armed armed bandit problem dates back to~\citep{agrawal1995continuum} and has been since extensively studied by~\citep{auer2007improved},~\citep{kleinberg2005nearly} and~\citep{kleinberg2019bandits} among other works. Although, continuum-armed bandit problems can be solved by discretizing the arm space such regret guarantees for such discretization schemes usually suffer the curse of dimensionality in terms of dimension of the action space. ~\citep{slivkins2011contextual} propose a policy that adaptively discretizes the joint context and arm space when the covariate arrivals are adversarial and rewards are scalar. Recent work by~\citep{russo2018satisficing} consider the problem of learning a near-optimal arm quickly for scalar rewards as opposed to the optimal arm over a large time horizon. Our work extends this line of work along two directions: we consider a vectorial mean-rewards as opposed to scalar rewards, and we assume that contexts arrive from a time-varying distribution as opposed to adversarial (worst-case arrival, as in~\citep{slivkins2011contextual}) or stochastic (i.i.d fixed-distribution arrival as in~\citep{perchet2013multi}). We further remark that as opposed to the non-stationary multi-armed bandit problem and variants thereof (see for eg.~\citep{besbes2014stochastic},~\citep{cheung2018hedging} and references therein) our work considers temporal variations in context distribution while assuming that the expected reward doesn't change with time. 

\item \textbf{Multi-Objective Learning:}  Learning under multiple-objectives has been mostly studied in the case of finite arms without contexts by~\citep{yahyaa2014annealing},~\citep{yahyaa2014knowledge} and~\citep{drugan2013designing}.~\citep{turugay2018multi} study a problem similar to the one considered in this paper, the multi-objective continuum-armed bandit problem with non-parametric expected rewards with adversarial covariate arrival. However, their policy is based on an adaptive discretization scheme which exploits similarity structure of the arm-context space under a different set of assumptions than those considered in this work.

\item \textbf{Covariate Shift:}
Covariate shift has been primarily studied in a classification setting wherein the marginal distribution of the covariates is different between source and target distributions. Policies for this class of problems are designed based on importance-sampling-based ideas such as~\citep{shimodaira2000improving} and~\citep{ben2007analysis}.~\citep{duchi2019distributionally} propose a distributionally robust model for learning under the distributional shift of the marginal covariate shift in an offline setting. Several other works such as~\citep{singh2021learning} (and the references therein) consider the problem of learning under covariate shift in the absence of bandit feedback. For the online setting,~\citep{si2020distributional} considers the problem of learning a distributionally robust policy from observational (bandit) data.~\citep{suk2020self} consider the problem of learning finite-armed contextual bandit model with scalar mean-rewards under covariate shift.
\end{enumerate}
\section{Properties of Pareto metric}
\label{sec:prop-pareto-metric}
In this section, we establish that $d_{p}$ as defined in~\eqref{eqn:pareto-metric} is a metric over $\pspace$, $\pspace$ is compact under $d_p$ and therefore $\pspace$ is complete. To this end, we derive several equivalent notions of dominance.
We first show the following gap property: $\Delta(X,k)=0$ if and only if $\mu_{k}(X) \in \pf{X}$. Proposition~\ref{prop:cpr} expresses the gap regarding a single point lying on the Pareto Front and its proof is immediate.  In the sequel, we will denote $\cP_1:=\pf{X_1}, \ \cP_2:=\pf{X_2},\ \ldots, \text{ for }X_1,X_2, \ldots \in \cspace$.

\begin{proposition}
\label{prop:cpr}
For a given preference cone $\cone$, we have:
\[
\Delta(k,\pf{X_2}) = \min_{k' \in \pf{X_2}}\max_{m \in [M]}\max\left\{0,\size{\log\frac{\mu^{m}_{k'}(X_1)}{\mu^{m}_{k}(X_2)}}\right\}
\]
where, $\log()$ is taken component-wise.
\end{proposition} 

\begin{lemma}[Preference metric $d_p$]
\label{lem:preference-metric}
$(\pspace,d_P)$ is a complete metric space. 
\end{lemma}
\begin{proof}
\begin{enumerate}
\item We first show that $d_p(\cP_1,\cP_2)$ is a metric. Let $\pf{X_1}, \pf{X_2} \in \pspace$. To show that $d_p$ is a metric, we show that: 
\begin{enumerate}
\item We now show that $d_{p}(\cP_1,\cP_{2}
) = 0 \iff \cP_{1} = \cP_{2}$. The implication $\cP_1 = \cP_{2} \implies d_{p}(\cP_1,\cP_{2})=0$ is immediate. For the other side, note that by Definition~\ref{defn:pareto-metric}, we have:
\begin{align*}
&& d_{p}\left(\cP_1,\cP_2\right) = 0  \\
&\implies& \max_{k \in \cP_{1}} \Delta(k,\cP_{2})=0 \ \text{ and } 
\max_{k \in \cP_{2}} \Delta(k,\cP_{1}) = 0 
\end{align*}
Further, $\max_{k \in \cP_{1}} \Delta(k,\cP_{2})=0$ implies:
\begin{align*}
\forall \ k \in \cP^{\ast}_{1}, k \notdom k', \ k' \in \cP_{2} \iff \forall \ k \in \cP_1, \ k \in \cP_2
\end{align*}
A similar argument using $\max_{k \in \cP_{1}} \Delta(k,\cP_1)=0$ implies that $\forall \ k \in \cP_{2}, \ k \notdom k', k' \in \cP^{\ast}_1$.
\item Symmetry: $d_{p}(\cP_1, \cP_2)$ is symmetric by definition (note that the gap in Definition~\ref{defn:pareto-gap} is not symmetric).
\item Triangle Inequality: 
\[
d_p(\mathcal{P}_1, \mathcal{P}_3) \le d_p(\mathcal{P}_1, \mathcal{P}_2) + d_p(\mathcal{P}_2, \mathcal{P}_3),
\]
where for any two sets \( \mathcal{P}, \mathcal{P}' \),
\[
d_p(\mathcal{P}, \mathcal{P}') := \max_{m \in [M]} \max \left\{ 
\max_{k \in \mathcal{P}} \min_{k' \in \mathcal{P}'} \left| \log \mu^m_{k'}(X') - \log \mu^m_k(X) \right|,
\max_{k' \in \mathcal{P}'} \min_{k \in \mathcal{P}} \left| \log \mu^m_k(X) - \log \mu^m_{k'}(X') \right|
\right\}.
\]
We analyze the first term inside the max:
\begin{align*}
\max_{m \in [M]} \max_{k \in \mathcal{P}_1} \min_{k' \in \mathcal{P}_3} 
\left| \log \mu^m_{k'}(X_3) - \log \mu^m_k(X_1) \right|.
\end{align*}
For any such \( k \in \mathcal{P}_1 \), fix an intermediate index \( k'' \in \mathcal{P}(X_2) \), and apply the triangle inequality:
\begin{align*}
\left| \log \mu^m_{k'}(X_3) - \log \mu^m_k(X_1) \right|
&= \left| \log \mu^m_{k'}(X_3) - \log \mu^m_{k''}(X_2) + \log \mu^m_{k''}(X_2) - \log \mu^m_k(X_1) \right| \\
&\le \left| \log \mu^m_{k'}(X_3) - \log \mu^m_{k''}(X_2) \right|
+ \left| \log \mu^m_{k''}(X_2) - \log \mu^m_k(X_1) \right|.
\end{align*}
Taking the inner minimum over \( k' \in \mathcal{P}_3 \), intermediate minimum over \( k'' \in \mathcal{P}(X_2) \), and outer maximum over \( k \in \mathcal{P}_1 \) and \( m \in [M] \). We get:
\begin{align*}
\max_{m \in [M]} \max_{k \in \mathcal{P}_1} \min_{k' \in \mathcal{P}_3} 
\left| \log \mu^m_{k'}(X_3) - \log \mu^m_k(X_1) \right|
&\le \max_{m \in [M]} \left(
\max_{k \in \mathcal{P}_1} \min_{k'' \in \mathcal{P}(X_2)} 
\left| \log \mu^m_{k''}(X_2) - \log \mu^m_k(X_1) \right| \right. \\
&\quad + \left.
\max_{k'' \in \mathcal{P}(X_2)} \min_{k' \in \mathcal{P}_3}
\left| \log \mu^m_{k'}(X_3) - \log \mu^m_{k''}(X_2) \right|
\right).
\end{align*}

Similarly, for the second term in the definition of \( d_p(\mathcal{P}_1, \mathcal{P}_3) \), we have:
\begin{align*}
\max_{m \in [M]} \max_{k \in \mathcal{P}_3} \min_{k' \in \mathcal{P}_1} 
\left| \log \mu^m_{k'}(X_1) - \log \mu^m_k(X_3) \right|
&\le \max_{m \in [M]} \left(
\max_{k \in \mathcal{P}_3} \min_{k'' \in \mathcal{P}(X_2)} 
\left| \log \mu^m_{k''}(X_2) - \log \mu^m_k(X_3) \right| \right. \\
&\quad + \left.
\max_{k'' \in \mathcal{P}(X_2)} \min_{k' \in \mathcal{P}_1}
\left| \log \mu^m_{k'}(X_1) - \log \mu^m_{k''}(X_2) \right|
\right).
\end{align*}

Putting it all together:
\begin{align*}
d_p(\mathcal{P}_1, \mathcal{P}_3) &\le \max \Big\{
\max_{m \in [M]} \Big[
\max_{k \in \mathcal{P}_1} \min_{k'' \in \mathcal{P}(X_2)} 
\left| \log \mu^m_{k''}(X_2) - \log \mu^m_k(X_1) \right| + 
\max_{k'' \in \mathcal{P}(X_2)} \min_{k' \in \mathcal{P}_3} 
\left| \log \mu^m_{k'}(X_3) - \log \mu^m_{k''}(X_2) \right| 
\Big], \\
&\quad\quad\ 
\max_{m \in [M]} \Big[
\max_{k \in \mathcal{P}_3} \min_{k'' \in \mathcal{P}(X_2)} 
\left| \log \mu^m_{k''}(X_2) - \log \mu^m_k(X_3) \right| + 
\max_{k'' \in \mathcal{P}(X_2)} \min_{k' \in \mathcal{P}_1} 
\left| \log \mu^m_{k'}(X_1) - \log \mu^m_{k''}(X_2) \right|
\Big]
\Big\}.
\end{align*}
This is equivalent to:
\[
d_p(\mathcal{P}_1, \mathcal{P}_2) + d_p(\mathcal{P}_2, \mathcal{P}_3),
\]

\end{enumerate}
\item We now show that $\pspace$ is compact under the metric $d_{p}$. Consider a sequence of Pareto fronts $\cP_1,\cP_2,\ldots,\cP_n \in \pspace$ and $\cP$ be the candidate for limiting Pareto front.
\begin{itemize}
\item Boundedness of $\cP$ is immediate. 
\item $\cP$ is convex since $\cP_n,\cP_{n+1}$ are convex and $\lambda \cP_{n} + (1-\lambda) \cP_{n+1}$ is also convex for all $\lambda \in [0,1]$. 
\item $\cP_n \rightarrow \cP, \ \text{therefore,} \ \forall \ \epsilon > 0, \exists \ N(\epsilon) \ \text{s.t.} \ \forall \ n > N(\epsilon)$ and $d_{p}(P_n,P) < \epsilon$. Let $\mu_{k}$ be a limit point of $\cP$, i.e., $\exists \ \text{a sequence} \ \mu_{k,n} \in \cP \ \text{such that} \ \mu_{k,n} \rightarrow \mu_{k}$. Since $d_{p}\left(\cP_n,\cP\right) \rightarrow 0$ for each $\mu_{k,n} \in \cP$ there exists $\mu_{k,n,m} \in \cP_n \ \text{s.t.} \ \mu_{k,n,m} \rightarrow \mu_{k,n}$. Using a diagonalization argument, we can obtain a subsequence $\mu_{k,n,m} \rightarrow \mu_{k}$. Since $\cP_n$ is compact, $\mu_k$ must lie in $\cP$ and therefore, $\cP$ is closed.
\end{itemize}
\end{enumerate}
\end{proof}

\section{Proof of Theorem~\ref{thm:cs-regret}}
\label{appendix:thm1}
\begin{definition}[Concentration of means]
\label{defn:concentration-arm-mean}
Given an arm $k \in [K]$ define the arm good-event $\gevent_{k}$ as the event that the learner is confident of the estimates of all the reward functions associated with the arm $k$ for all levels $h \in \mN$:
\[
\gevent_{k}=\bigcap_{h \in \mN}\bigcap_{m \in [\numobj]}\Big\{\sup_{x \in \cX}  \big\vert \mu^{(m)}_{k}-\hat{\mu}^{(m)}_{k,t} \big\vert \in [\lowktl{k}{t}{m},\upktl{k}{t}{m}] \Big\},
\]
where, $L_{k,t}^{(m)} = c_{1}V^{\beta}_h - c_{2}\sqrt{\frac{\log\left(\frac{KL}{\delta}\right)}{\numkt\binid}} $ and $U_{k,t}^{(m)} = c_{1}V^{\beta}_h + c_{2}\sqrt{\frac{\log\left(\frac{KL}{\delta}\right)}{\numkt\binid}}$. 
\end{definition}

\begin{lemma}[Concentration]
\label{lem:arm-concentration}
For all $t \in \mN, k \in [K], m \in [\numobj]$ and $\binid$, we have: 
\begin{equation*}
\sup_{X \in \cX} \Big\vert \hat{\mu}_{k,t}(X) - \mu_{k,t}(X)\cdot \ind{X \in \bin{h}{i}} \Big\vert \le c_{1}V^{\beta}_h + c_{2}\sqrt{\frac{\log\left(\frac{KL}{\delta}\right)}{\numkt\binid}}
\end{equation*}
holds with probability $1-\delta$. 
\end{lemma}

\begin{proof}
From~\eqref{eq:muhat-hi-def}, we have that: 
\begin{eqnarray*}
\muestm{m}{k}{t} = \frac{1}{\numkt\binid} \sum_{s=1}^{t} r^{m}_{s} 
\end{eqnarray*}
Define the expected reward 
\[
\bar{\mu}_{k,t}\binid = \mE\left[\muestm{m}{k}{t} \vert \cF_{t-1}\right]
\]
From triangle inequality, we have: 
\[
\big\vert \muestm{m}{k}{t} - \mu_{k} \big\vert \le \big\vert \muestm{m}{k}{t}-\bar{\mu}_{k,t}\big\vert + \big\vert \bar{\mu}_{k,t}-\mu_{k} \big\vert
\] 
By Assumption~\ref{assmpt:holder_cont}, we have:
\[
\big\vert \bar{\mu}_{k,t}-\mu_{k} \big\vert \le V_h^{\beta}
\]
For the other term, by Hoeffding's inequality, we have: 
\[
\big\vert \muestm{m}{k}{t} - \bar{\mu}_{k,t} \big\vert \le \sqrt{\frac{\log\frac{2}{\delta}}{\numkt\binid}}
\]
Taking a union bound over $\armset$ and $[\numobj]$, we have with probability $1-\delta$:
\[
\big\vert \muestm{m}{k}{t} - \mu_{k} \big\vert \le \sqrt{\frac{\log\frac{2KL}{\delta}}{\numkt\binid}}
\]
\end{proof}

\begin{lemma}[Tree depths are monotonic]
\label{lem:monotonic-levels}
If bin $\bin{h}{i}$ was selected at time $t'$ then descendants $\bin{h}{i}$ are selected at $t>t'$.
\end{lemma}

\begin{proof}
We first show that the depth selection mechanism doesn't skip any bins. For the sake of contradiction, suppose a descendant bin $\bin{h_s}{i_s} \in \text{descendant}(\bin{h_t}{i_t})$ was selected before $\bin{h_t}{i_t}$ at some time $s < t$. Then by bin expansion criteria (Line~\ref{algo-line:bandwidth-selection} in Algorithm~\ref{algo:cpf}), we have that:
\begin{equation*}
V^{\beta}_{h_t} \ge \sqrt{\frac{8K\log\left(\frac{KL}{\delta}\right)}{n_{h_t,t}}} \implies n_{h_t,t} \ge \frac{8K\log\left(\frac{KL}{\delta}\right)}{V^{2\beta}_{h_t}}
\end{equation*}
Since $V_{h_t} > 2V_{h_s}$, we have
\begin{equation*}
\frac{8K\log\left(\frac{KL}{\delta}\right)}{V^{2\beta}_{h_t}} \le \frac{8K\log\left(\frac{KL}{\delta}\right)}{4V^{2\beta}_{h_s}} \le \frac{n^{2}_{h_s,s}}{4} \le n_{h_s,s}
\end{equation*}
Therefore, for some $s' < s$, such that:
\begin{eqnarray*}
n_{h_s,s'} \ge \frac{8K\log\left(\frac{KL}{\delta}\right)}{V^{2\beta}_{h_t}} \implies V^{2\beta}_{h_t} \ge \frac{8K\log\left(\frac{KL}{\delta}\right)}{n_{h_s,s'}} \ge \frac{8K\log\left(\frac{KL}{\delta}\right)}{n_{h_t}} 
\end{eqnarray*} 
Therefore, $V_{h_s'} \le V_{h_t}$ leading to a contradiction. 
\end{proof}

\begin{lemma}[Arm and covariate counts]
\label{lem:arm-covariate-count}
Suppose bin $\bin{h}{i}$ was selected at time $t$, then with probability $1-\delta, \ \forall \ k \in \activearm$,  
\[
\numkt(h_t,i_t) \ge \frac{n_{h^{\ast}_t}}{4K}
\]
\end{lemma}

\begin{proof}
This proof is inspired from Lemma~3 in~\citep{suk2020self}. Fix an arm $i \in \activearm$. By~\eqref{eq:n-def}, we have:
\[
\numkt{(h_t,i_t)} = \sum_{s=1}^{t} \ind{X_s \in \bin{h_s}{i_s}}
\]
For $h=0$, the tree contains only the root. Therefore, $n_{h_t,t} = K\log(\frac{KL}{\delta}) +1$. For each round so far we have pulled an arm uniformly at random (Line~\ref{algo-line:uniform-arm-play} in Algorithm~\ref{algo:cpf}), therefore, 
\[
\mE\left[\numkt(h_t,i_t)\right]  \ge 8K\log (\frac{KL}{\delta}) 
\]
Using Chernoff's inequality, we have: 
\[
\source\left(\numkt(h_t,i_t) \le \frac{\mE\numkt(h_t,i_t)}{2}\right) \le \frac{\delta}{K}
\]
For $h \ge 1$, the tree deepens and the arms pulls are no longer independent. Therefore, we cannot use the above argument. To circumvent this issue, we create a coupling between the arm pulls of Algorithm~\ref{algo:cpf} and an independently randomized (fictitious) set of arm pulls.
\medskip 

Let $t'$ denote the first time $\text{Pa}\left({\bin{h}{i}}\right)$ is visited. By Lemma~\ref{lem:monotonic-levels}, we have that $t'<t$. By active arm refinement rule, Line~\ref{algo-line:active-arm-refine} in Algorithm~\ref{algo:cpf}, we have that $\activearm \subseteq \cA_{(h_{t'},i_{t'})}$. Let $n_{k,[t',t]}\binid$ denote the number of times arm $k$ is played in bin $\binid$ in the interval $[t',t]$ by Algorithm~\ref{algo:cpf} and let $\tilde{n}_{k,[t',t]}\binid$ denote draws from $\text{Bin}\left(n_{[t,t_0]}(h),\frac{1}{\cA_{(h_{t'},i_{t'})}}\right)$. Since $\vert \cA_{(h_t,i_t)} \vert \le \vert \cA_{(h_{t'},i_{t'})} \vert$, we have that $n_{k,[t',t]} \ge \tilde{n}_{k,[t',t]}$. 

\medskip 
Since $\bin{h}{i} \in \leaft$, we have: 
\begin{equation}
\label{eqn:bin-play}
V^{\beta}_{h_t} \ge \sqrt{\frac{\log\left(\frac{KL}{\delta}\right)}{n_{h_t,t}}} \implies  n_{h_t,t} \ge \frac{K\log\left(\frac{KL}{\delta}\right)}{V^{2\beta}_{h_t}}
\end{equation}
Similarly, since $t'$ is the first time $\text{Pa}\left(\bin{h}{i}\right)$ is chosen, 
\begin{equation}
\label{eqn:parent-bin-play}
\sqrt{\frac{8K\log(\frac{KL}{\delta})}{n_{2h_t,t'}}} > V^{\beta}_{h_t} \implies n_{2h_t,t'} \ge \frac{8K\log(\frac{KL}{\delta})}{V^{2\beta}_{h_t}}
\end{equation}
From the last two equations, we have: 
\[
n_{h_t,t} - n_{[t',t]}\binid \le n_{2h_t,t'} \le \frac{n_{h_t}}{2}
\]
This implies that: $n_{[t',t]}\binid \ge \frac{n_{h_t}}{2}$. For every $s \in [t',t]$ we pull arm $k$ w.p. $\frac{1}{K}$ and thus: 
\[
\mE\left[\tilde{n}_{k,[t',t]}\binid\right] \ge \frac{n_{[t',t]}\binid}{K} \ge \frac{n_{h_t}}{K} 
\]
From equation~\eqref{eqn:bin-play}, we have that: 
\[
\frac{8K\log\left(\frac{KL}{\delta}\right)}{V^{2\beta}_{h_t}} \ge 8K\log\left(\frac{KL}{\delta}\right)
\]
Since $\tilde{n}_{k,[t',t]}\binid$ is sampled independently from Binomial distribution using a Chernoff bound, we have: 
\[
\source\left(\tilde{n}_{k,[t',t]}\binid \le \frac{n_t(h_t)}{K}\right) \le \source\left(\tilde{n}_{k,[t',t]}\binid \le \frac{\mE\left[\tilde{n}_{k,[t',t]}\binid\right]}{2} \right) \le \frac{\delta}{K}
\]
Therefore, with probability $1-\frac{\delta}{K}$ we have that:
\[
n_{k,t}\binid \ge n_{k,[t',t]}\binid \ge \tilde{n}_{k,[t',t]}\binid \ge \frac{n_t(h_t)}{K} 
\]
\end{proof}

\begin{lemma}[Refining active arms]
For any $t \le T, \ \text{such that}\ X_t \in \bin{h_t}{i_t}, \ \pf{X_t} \subset \cA_{(h_t,i_t)}$ with probability $1-\delta$ and $C_{h_t,t}$ as specified in Lemma~\ref{lem:arm-concentration}.
\end{lemma}

\begin{proof}
We establish this by contradiction. Suppose, $k \in \pf{X_s}$ but $k \notin \activearm$. Then, 
\begin{eqnarray*}
\mu_{k}(X_s) &\stackrel{(b)}{\notdom}& \mu_{k'}(X_s) \ \forall \ k' \in \activearm \\
\hat{\mu}_{k',t} - C_{t,h_t} &\stackrel{(c)}{\prec}& \hat{\mu}_{k,t} + C_{h_t,t} \\
\vert \hat{\mu}_{k',t} - \hat{\mu}_{k,t} \vert &\weakdom& C_{h_t,t}
\end{eqnarray*}
where, $(a)$ follows since $k \in \pftrue$, $(b)$ follows since $\activearm \subseteq \armset$, $(c)$ follows by Lemma~\ref{lem:arm-concentration}. The last inequality establishes the contradiction.
\end{proof}

\begin{lemma}[Regret under Good Event]
\label{lem:regret-good-event}
We have that: 
\begin{eqnarray*}
\mE\left[\sum_{t=t_{p}+1}^{T} \pdist{\pfestm}{\pftrue(X_t)} \big\vert \cH_t \right] &\le&  c_{9} \left(K\log\left(\frac{KL}{\delta}\right)\max\left\{\frac{\rho(\source,\target)}{t_p},\frac{\rho(\target,\target)}{\tau}\right\}\right)^{\left(\frac{(\alpha+1)}{\alpha}\right)\left(\frac{\beta+1}{\beta}\right)} \left(\frac{K\log\frac{KL}{\delta}}{\tau}\right)^{\frac{1}{\alpha}} \\
&+& c_{10} \left(\frac{K \log\left(\frac{KL}{\delta}\right)}{\min\{t_p,\tau\}}\right)^{\frac{\beta(\alpha+1)}{\beta}}
\end{eqnarray*}
\end{lemma}

\begin{proof}
We decompose the regret as follows: 
\begin{eqnarray*}
\regt &=& \mE\left[\sum_{t=t_{p}+1}^{T} \pdist{\pfestm}{\pftrue(X_t)}\right] \\
&\stackrel{(a)}{=}& \mE\left[\sum_{t=t_{p}+1}^{T}\left(\pdist{\pfestm}{\pftrue(X_{(h_t,i_t)})} + \pdist{\pftrue(X_{(h_t,i_t)})}{\pftrue(X_t)}\right)\right] \\
&=& \underbrace{\mE\left[\sum_{t=t_{p}+1}^{T} \pdist{\pfestm}{\pftrue(X_{(h_t,i_t)})} \right]}_{\text{Term-I}} \\
&+& \underbrace{\mE\left[\sum_{t=t_{p}+1}^{T} \pdist{\pftrue(X_{(h_t,i_t})}{\pftrue(X_t)}\right]}_{\text{Term-II}}
\end{eqnarray*}
where, $(a)$ follows from triangle inequality and the fact that $\pdist{\cdot}{\cdot}$ is a metric (Lemma~\ref{lem:preference-metric}).  
\medskip

\textbf{Term-I}:
\small{
\begin{align*}
\pdist{\pfestm}{\pftrue(X_{(h_t,i_t)})}
&\le \max\Bigg\{ 
\min_{k \in \pfestm} \max_{k' \in \pftrue} \left| \log \mu_{k'}(X_{(h_t,i_t)}) - \log \hat{\mu}_{k,t}(X_t) \right|, \\
&\hspace{1.6em}
\min_{k \in \pftrue} \max_{k' \in \pfestm} \left| \log \hat{\mu}_{k',t}(X_t) - \log \mu_k(X_{(h_t,i_t)}) \right| 
\Bigg\}
\end{align*}
}
We note that:
\begin{equation}
\label{eqn:log-bound}
\log a - \log b \le \frac{1}{\min\{a,b\}} \vert a-b \vert 
\end{equation}
~\eqref{eqn:log-bound} can be established using the mean-value theorem. To see this note that $\log x$ is differentiable in $(0,\infty)$. By mean-value theorem, for some $c \in (a,b)$ we have: 
\[
\log a  - \log b \le \frac{1}{c} \size{a-b} \le \frac{1}{\min\{a,b\}}\size{a-b}
\]
Using~\eqref{eqn:log-bound}, we have for all $k,X$ and $m$:
\[
\size{\log \mu^{m}_{k}(X) - \log \mu^{m}_{k'}(X)} \le c_{1}\size{\mu^{m}_{k}(X) - \mu^{m}_{k'}(X)}
\]
Let 
\[
A_t = c_{1}\min_{k \in \pfestm} \max_{k' \in \pftrue} \max_{m}\size{\mu_{k'}(X_{(h_t,i_t)}) - \hat{\mu}_{k,t}(X_t)}, \ B_t = c_{1}\min_{k \in \pftrue} \max_{k' \in \pfestm} \max_{m}\size{\hat{\mu}_{k',t}(X_t) - \mu_{k}(X_{(h_t,i_t)})}
\]
implying:
\begin{eqnarray*}
\pdist{\pfestm}{\pftrue(X_{(h_t,i_t)})} &\le& \max\left\{ A_t, B_t \right\} \\
&\le& A_t + B_t + \vert A_t - B_t \vert
\end{eqnarray*}

Hence, Term-I is: 
\begin{eqnarray*}
\pdist{\pfestm}{\pftrue(X_{(h_t,i_t)})} &\le& \frac{1}{2}\mE\left[\sum_{t=t_p+1}^{T} \left(A_t+B_t + \vert A_t - B_t \vert\right) \ind{X_t \in \bin{h_t}{i_t}} \ind{\bin{h_t}{i_t}\in \leaft} \right] 
\end{eqnarray*}
Focusing on $A_t$, with probability $1-\delta$, we have:
\begin{eqnarray*}
A_t  &=& \min_{k \in \pfestm}\max_{k' \in \pftrue} \mutrue{m}{k'}(X_{(h_t,i_t)}) - \muestm{m}{k}{t}(X_t)  \\
&=&\min_{k \in \hat{P}_t}\max_{k' \in \pftrue} \mutrue{m}{k'}(X_{(h_t,i_t)}) - \mutrue{m}{k}(X_{(h_t,i_t)}) + \mutrue{m}{k}(X_{h_t,i_t}) -\muestm{m}{k}{t}(X_t) \\
&\le&  \underbrace{\max_{k' \in \cP^{\ast}} \mutrue{m}{k'}(X_{(h_t,i_t)})-\mutrue{m}{k}(X_{(h_t,i_t)})}_{A_{1t}} + \underbrace{\max_{X \in \bin{h_t}{i_t}} \Big\vert \mutrue{m}{k}(X) -\muestm{m}{k}{t}(X)\Big\vert}_{A_{2t}}
\end{eqnarray*}
We bound the first term as:
\begin{eqnarray*}
&& \mE\left[\sum_{t=t_p+1}^{T}\max_{k' \in \pftrue}\left( \mutrue{m}{k'}(X_{(h_t,i_t)})-\mutrue{m}{k}(X_{(h_t,i_t)})\right) \right] \\
&\stackrel{(a)}{\le}&\mE\left[\sum_{t=t_p+1}^{T}A_{1t} \cdot \ind{0 \le  A_{1t} \le V_{h_t} }\right] \\
&\stackrel{(b)}{\le}&\mE\left[\sum_{t=t_p+1}^{T}A_{1t} \ind{0 \le A_{1t} \le V^{\beta}_{h^{\ast}_t} + s_{h^{\ast}_t}} \right] \\
&\stackrel{(c)}{\le}& \sum_{t=t_p+1} \mE\left[c_{4} V_{h^{\ast}_t}^{\beta}\cdot \ind{0 \le A_{1t} \le  V_{h^{\ast}_t}}+\sqrt{\frac{\log\left(\frac{KL}{\delta}\right)}{\max\{t_{p}\source\left(\cT_{h^{\ast}_t}\right),\tau \target\left(\cT_{h^{\ast}_t}\right)\}}}\cdot\ind{0 \le A_{1t} \le s_{h^{\ast}_t}}\right] 
\end{eqnarray*}
where, $(a)$ and $(b)$ follow from Lemma~\ref{lem:oracle-level} and $(c)$ follows from $ \ind{x \le u+v} \le \ind{x \le 2u} + \ind{x \le 2v}$. In order to bound term $A_{2t}$, we use Lemma~\ref{lem:arm-concentration} which gives a high-probability bound and convert it into a bound in expectation as follows: 
\begin{eqnarray*}
&& \mE\left[\sum_{t=t_p+1}^{T}\max_{X \in \bin{h_t}{i_t}}\left( \mutrue{m}{k'}(X)-\muestm{m}{k}{t}(X)\right)\right] \\
&=&\sum_{t=t_p+1}^{T} \frac{1}{c}\left[\left(c_1V^{\beta}_{h_t}+c_2\sqrt{\frac{K\log\frac{KL}{\delta}}{n_{k,t}}}\right)+\delta\left(u-c_1V^{\beta}_{h_t}-c_2\sqrt{\frac{K\log\frac{KL}{\delta}}{n_{k,t}}}\right)\right]  \\
&\le& 
\end{eqnarray*}
From the previous two panels, we have:
\begin{eqnarray*}
\mE\left[\sum_{t=t_p+1}^{T} A_t \right] &\le& c_{9} \left(K\log\left(\frac{KL}{\delta}\right)\max\left\{\frac{\rho(\source,\target)}{t_p},\frac{\rho(\target,\target)}{\tau}\right\}\right)^{\left(\frac{(\alpha+1)}{\alpha}\right)\left(\frac{\beta+1}{\beta}\right)} \left(\frac{K\log\frac{KL}{\delta}}{\tau}\right)^{\frac{1}{\alpha}} \\
&+& c_{10} \left(\frac{K \log\left(\frac{KL}{\delta}\right)}{\min\{t_p,\tau\}}\right)^{\frac{\beta(\alpha+1)}{\beta}}
\end{eqnarray*}
A similar argument shows that: 
\begin{eqnarray*}
\mE\left[\sum_{t=t_p+1}^{T} B_t \right] &\le& c_{9} \left(K\log\left(\frac{KL}{\delta}\right)\max\left\{\frac{\rho(\source,\target)}{t_p},\frac{\rho(\target,\target)}{\tau}\right\}\right)^{\left(\frac{(\alpha+1)}{\alpha}\right)\left(\frac{\beta+1}{\beta}\right)} \left(\frac{K\log\frac{KL}{\delta}}{\tau}\right)^{\frac{1}{\alpha}} \\
&+& c_{10} \left(\frac{K \log\left(\frac{KL}{\delta}\right)}{\min\{t_p,\tau\}}\right)^{\frac{\beta(\alpha+1)}{\beta}} 
\end{eqnarray*}
and:
\begin{eqnarray*}
&& \mE\left[\sum_{t=t_p+1}^{T} \vert A_t - B_t \vert \right] \\
&\le& \mE\left[ \sum_{t=t_p+1}^{T} A_t \right] + \mE\left[ \sum_{t=t_p+1}^{T} B_t \right] \\
&\le& 2c_{9} \left(K\log\left(\frac{KL}{\delta}\right)\max\left\{\frac{\rho(\source,\target)}{t_p},\frac{\rho(\target,\target)}{\tau}\right\}\right)^{\left(\frac{(\alpha+1)}{\alpha}\right)\left(\frac{\beta+1}{\beta}\right)} \left(\frac{K\log\frac{KL}{\delta}}{\tau}\right)^{\frac{1}{\alpha}} \\
&+& 2c_{10} \left(\frac{K \log\left(\frac{KL}{\delta}\right)}{\min\{t_p,\tau\}}\right)^{\frac{\beta(\alpha+1)}{\beta}} 
\end{eqnarray*}
\textbf{Term-II:}
\medskip 
From the margin condition, we have that:
\small{
\begin{eqnarray*}
&& \mE\left[\sum_{t=t_{p}+1}^{T} \pdist{\pftrue(X_{(h_t,i_t})}{\pftrue(X_t)}\right] \\
&\le& \mE\left[\sum_{t=t_p+1}^{T}\max\left\{ \min_{k \in \pf{X_t}} \max_{k' \in \pf{X_{(h_t,i_t)}}} \mu_{k'}(X_{(h_t,i_t)}) -\mu_{k}(X_t) , \min_{k \in \pf{X_{(h_t,i_t)}}} \max_{k' \in \pf{X_t}} \mu_{k'}(X_t) - \mu_{k}(X_{(h_t,i_t)}) \right\}\right] \\
&\le& \mE\left[\sum_{t=t_p+1}^{T} 2V^{(1+\alpha)}_{h_t} \right]\\
&\le&  c_{9} \left(K\log\left(\frac{KL}{\delta}\right)\max\left\{\frac{\rho(\source,\target)}{t_p},\frac{\rho(\target,\target)}{\tau}\right\}\right)^{\left(\frac{(\alpha+1)}{\alpha}\right)\left(\frac{\beta+1}{\beta}\right)} \left(\frac{K\log\frac{KL}{\delta}}{\tau}\right)^{\frac{1}{\alpha}} \\
&+& c_{10} \left(\frac{K \log\left(\frac{KL}{\delta}\right)}{\min\{t_p,\tau\}}\right)^{\frac{\beta(\alpha+1)}{\beta}}
\end{eqnarray*}
}
\end{proof}

Recall that by Line~\ref{algo-line:bandwidth-selection} in Algorithm~\ref{algo:cpf} the level of the regressogram is selected to balance the bias and variance associated with the estimator. Define the optimal regression rate as: $h^{\ast}_t = \arg\min_{h \in \mN} c_{1} \sqrt{\frac{\log(\frac{KL}{\delta})}{n_{h}}} + c_{2} V^{\beta}_{h}$. In Lemma~\ref{lem:oracle-level}, we show that this doesn't incur an error larger than the oracle level for this bandit problem
\begin{lemma}
\label{lem:oracle-level}
For any $1 \le s \le T$ with probability $1-$, wrt conditional distribution of we have that: 
\begin{equation*}
V^{\beta}_{h_t} \le \Psi(h_t) \le \Psi(h^{\ast}_t) \le c_{1} V^{\beta}_{h^{\ast}_t} + c_{2}\sqrt{\frac{2\log\left(\frac{KL}{\delta}\right)}{{\max\left\{t_p \source(\cT_{h^{\ast}_t}),\tau \target(\cT_{h^{\ast}_t})\right\}}}} 
\end{equation*}
\end{lemma}

\begin{proof}
We have that:
\[
\mE\left[n_{h^{\ast}_t}\right] = t_{p} \source(\cT_{h^{\ast}_{t}}) + \tau \target(\cT_{h^{\ast}_{t}}) \ge \max\{t_{p} \source(\cT_{h^{\ast}_t}),\tau \target(\cT_{h^{\ast}_t})\}
\]
Then, by a Chernoff bound, we have that:
\[
\source\left( n_{h^{\ast}_t} \le \frac{1}{2} \mE\left[n_{h^{\ast}_t}\right] \right) \le \exp\left(\frac{-1}{8}\mE\left[n_{h^{\ast}_t}\right]\right) \le \delta
\]
Then, with probability at least $1-\delta$:
\[
\Psi_t(h^{\ast}_t)\le \left(c_1V^{\beta}_{h^{\ast}_t}+c_2\sqrt{\frac{2\log\left(\frac{KL}{\delta}\right)}{\max\left\{t_p \source(\cT_{h^{\ast}_t}),\tau \target(\cT_{h^{\ast}_t})\right\}}}\right)
\]
\end{proof}

For the remainder of this proof, let $s_{h^{\ast}_t} = c_2\sqrt{\frac{2\log\left(\frac{KL}{\delta}\right)}{\max\left\{t_p \source(\cT_{h^{\ast}_t}),\tau \target(\cT_{h^{\ast}_t})\right\}}}$. Lemma~\ref{lem:oracle-level}, allows us to quantify $A_{1t}$ (see Lemma~\ref{lem:regret-good-event}) in terms of the estimation error when using the oracle level. We now bound this estimation error using the margin condition and dissimilarity metric in Definition~\ref{defn:dissimilarity-metric}. We define the following concentration events used until now:
\begin{equation}
\label{eqn:covariate-counts}
\cN_t = \left\{ \max\{t_p \source(\cT_{h^{\ast}_t}),  \tau \target(\cT_{h^{\ast}_t})\} \ge  \log\left(\frac{KL}{\delta}\right) \right\}
\end{equation}
In the light of the above results, define the concentration: 
\[
\cM_t= \Big\{n_{h^{\ast}_t} \ge \mE\left[ n_{h^{\ast}_t}\right]\Big\}
\]
So far we have considered three events: 
\begin{enumerate}
\item $\cG_t$: True mean lies within the confidence ball (Lemma~\ref{lem:arm-concentration})
\item $\cN_t$: Number of covariates in each bin are sufficiently large
\item $\cM_t$: Sufficient covariates in the optimal-regression bin
\end{enumerate}
Let $\cH_t = \cG_t \cap \cN_t \cap \cM_t$. Then under $\cH_t$, Lemma~\ref{lem:arm-concentration} and Lemma~\ref{lem:oracle-level} (and other consequent results) hold. For the first term, using the margin condition (Assumption~\ref{assmpt:margin}), we have:
\begin{equation}
\label{eqn:bias-bound}
\mE_{Q} \left[V^{\beta}_{h^{\ast}_t}\cdot\ind{0 \le \mutrue{m}{k} - \mutrue{m}{k'}  \le V_{h^{\ast}_t}}\right] \le \left(V_{h^{\ast}_t}\right)^{\beta(1+\alpha)}
\end{equation}
$s_{h^{\ast}_t}$ depends on $\source$ and $\target$ and therefore, a more delicate analysis is required. To this end, we focus on lower bounding $d_t = \sqrt{\max\left\{t_p \source(\cT_{h^{\ast}_t}),\tau \target(\cT_{h^{\ast}_t})\right\}}$. This is a combination of two terms which we bound separately. Define $d_{1t} = \left(\target(\cT_{h_t^{\ast}})\right)^{-1/2}$ and $d_{2t} = \left(\source(\cT_{h_t^{\ast}})\right)^{-1/2}$ For some $\eta$ (to be determined later) consider the decomposition: 
\[
\mE[s^{\ast}_t] = \mE[s^{\ast}_t \ind{0 \le A_{1t}  \le s^{\ast}_t} \left(\ind{ d_t \ge \eta} + \ind{d_t < \eta}\right)]
\]
When $d_t \le \eta$, we have 
\begin{equation}
\label{eqn:q-t-lower}
\mE[s^{\ast}_t\ind{0 \le A_{1t} \le s^{\ast}_t}\ind{d_{1t} \le \eta}] \le \sqrt{\frac{\log\frac{KL}{\delta}}{\left(T-t_p+1\right)}} \mE[d_t\ind{0 \le A_{1t} \le }\ind{d_t < \eta}] \le \eta^{-(1+\alpha)} \sqrt{\frac{\log\frac{KL}{\delta}}{\tau}}
\end{equation}
where, the last inequality follows from the margin assumption~\ref{assmpt:margin}. For the other case $d_t \ge \eta$, we have: 
\begin{equation}
\label{eqn:q-t-upper}
\mE\left[s^{\ast}_t\ind{0 \le A_{1t} \le s^{\ast}_t}\ind{d_{1t}\ge \eta}\right] \le \mE\left[ d_{1t} \ind{d_{1t} \ge \eta}\right]\sqrt{\frac{\log\left(\frac{KL}{\delta}\right)}{\tau}} 
\end{equation}
\begin{lemma}
For any $h$, we have: 
\begin{eqnarray}
\mE_{Q}\left[ d_{1t} \cdot \ind{ d_{1t} \ge \eta} \right]&\le& c_5 \frac{\rho_{h^{\ast}_t}(\target,\target)}{\eta} \label{eqn:d1t-q}\\
\mE_{Q}\left[ d_{2t} \cdot \ind{d_{2t} \ge \eta}\right]&\le& c_6\frac{\rho_{h^{\ast}_t}(\source,\target)}{\eta} \label{eqn:d1t-p} 
\end{eqnarray}
\end{lemma}

\begin{proof}
Using tail probability formula: 
\[
\mE_{Q}[d_{1t} \ind{d_t \ge \eta}] = \int_{0}^{\infty}\left( d_{1t} \cdot \ind{d_{1t} \ge \eta} \ge \xi \right) d\xi
\]
Noting that $d_{1t} \cdot \ind{d_{1t} \ge \eta} \ge \xi \iff d_{1t} \ge \eta \wedge \xi$. This gives us: 
\begin{eqnarray*}
\int_{0}^{\infty} \target \left( d_{1t} \ind{d_{1t} \ge \eta} \ge \xi \right) ds &=& \int_{0}^{\eta} \target\left( d_{1t} \ge \eta \right) d\eta+ \int_{\eta}^{\infty} \target\left( d_{1t} \ge \xi \right) d\xi\\
&\le& \int_{\eta}^{\infty} \frac{\mE_{Q}(d^{2}_{1t})}{\eta^{2}} + \frac{\mE_{Q}(d^{2}_{1t})}{\eta^{2}} 
\end{eqnarray*}
where, the last inequality follows from Chebyshev's inequality. To bound the variance, we have: 
\[
\mE_{Q} \left[d^{2}_{1t}\right] = \mE_{\target}\left(\frac{1}{\target(\cT_{h^{\ast}_t})}\right) \le  \frac{1}{}\mE_{Q}\left(\frac{1}{\target(\cT_{h^{\ast}_t})}\right) \le c_{5}\rho_{h^{\ast}_t}(\target,\target)
\]
Therefore, we get: 
\[
\frac{\mE_{Q}\left[d^{2}_{1t}\right]}{\eta} \le \frac{c_5\rho_{h^{\ast}_t}(\target,\target)}{\eta}
\]
Using the definition of Dissimilarity metric (Definition~\ref{defn:dissimilarity-metric}) the analogous bound for $P$ is given by:
\begin{equation}
\label{eqn:dissimilar-p}
\frac{\mE_{\target}\left[d^{2}_{2t}\right]}{\eta} =  \left[\frac{1}{\source(\cT_{{h^{\ast}_t}})} \right] \le \frac{1}{\eta}\mE_{\target}\left[\frac{1}{\source(\cT_{h^{\ast}_t})}\right] \le \frac{\rho_{h^{\ast}_t}(\source,\target)}{\eta} 
\end{equation}
\end{proof}
Therefore, 
\begin{equation}
\label{eqn:q-t-upper}
\mE[d_{1t} \ind{d_{1t} \ge \eta} \ind{0 \le A_{1t} \le s^{\ast}_t}] \le \mE_{Q}\left[ d_{1t} \ind{d_{1t} \ge \eta} \right] \le c_5\sqrt{\frac{\log\left(\frac{KL}{\delta}\right)}{\tau}} \frac{\rho_{h^{\ast}_t}(\target,\target)}{\eta}
\end{equation}
Using~\eqref{eqn:q-t-lower}, we have:
\begin{eqnarray*}
\mE\left[s^{\ast}_t\ind{0 \le A_{1t} \le s^{\ast}_t} \right] \le \eta^{-(1+\alpha)} \sqrt{\frac{\log\frac{KL}{\delta}}{\tau}} + c_5\max\left\{\sqrt{\frac{\log\frac{KL}{\delta}}{\tau}} \frac{\rho_{h^{\ast}_t}(\target,\target)}{\eta},\sqrt{\frac{\log\frac{KL}{\delta}}{t_p}} \frac{\rho_{h^{\ast}_t}(\source,\target)}{\eta}\right\}
\end{eqnarray*}
In order to balance the two terms we set $\eta = c_7\left(\frac{K\log\frac{KL}{\delta}}{\max\left\{\rho_{h^{\ast}_t}(Q,Q)\sqrt{\frac{K\log\frac{KL}{\delta}}{\tau}},\ \rho_{h^{\ast}_t}(P,Q)\sqrt{\frac{K\log(\frac{KL}{\delta})}{t_p}}\right\}}\right)^{\frac{-1}{\alpha}}$. From Lemma~\ref{lem:oracle-level}, $V_{h^{\ast}_t} = \left(\frac{K\log\frac{KL}{\delta}}{t}\right)^{\frac{1}{\beta}}$ which implies:
\begin{eqnarray}
\label{eqn:st-a1t-final}
\mE\left[ s^{\ast}_t \ind{0 \le A_{1t} \le s^{\ast}_t} \right] \le c_{9} \max\left\{\rho_{h^{\ast}_t}(\source,\target)\frac{K\log\frac{KL}{\delta}}{t_p},\rho_{h^{\ast}_t}(\target,\target)\frac{K\log\frac{KL}{\delta}}{\tau}\right\}^{\left(\frac{(\alpha+1)}{\alpha}\right)\left(\frac{\beta+1}{\beta}\right)} \left(\frac{K\log\frac{KL}{\delta}}{\tau}\right)^{\frac{1}{\alpha}}.
\end{eqnarray}
Combining this with equation~\eqref{eqn:bias-bound}, we get: 
\small{
\begin{align}
\mE\left[\Psi(h^{\ast}_t)\right] \le c_{9} \left(K\log\left(\frac{KL}{\delta}\right)\max\left\{\frac{\rho_{h^{\ast}_t}(\source,\target)}{t_p},\frac{\rho_{h^{\ast}_t}(\target,\target)}{\tau}\right\}\right)^{\left(\frac{(\alpha+1)}{\alpha}\right)\left(\frac{\beta+1}{\beta}\right)} \left(\frac{K\log\frac{KL}{\delta}}{\tau}\right)^{\frac{1}{\alpha}} + c_{10} \left(\frac{K \log\left(\frac{KL}{\delta}\right)}{\min\{t_p,\tau\}}\right)^{\frac{\beta(\alpha+1)}{\beta}}
\end{align}
}
Now, we bound the regret occurring under a bad event, $\ind{\overline{\cH}_t}$.  By definition of $\cH_t$, we have that: 
\[
\Pr\left(\overline{\cH}_t\right) \le \Pr\left(\overline{\cN}_t\right) + \Pr\left(\overline{\cG}_t\right) + \Pr\left(\overline{\cM}_t\right)
\]
We bound each term individually as follows:
\begin{itemize}
\item
From Lemma~\ref{lem:arm-concentration}, under the event $\overline{\cG}$, the regret is $1$ and it happens with probability $\delta$. Therefore, its contribution to regret is given by: $T\delta$. 
\item 
From Lemma~\ref{lem:oracle-level}, the event $\overline{\cN}_t$, occurs with probability $\delta$
\item 
Under the event $\cM_t$, we have that: 
\begin{eqnarray*}
\Pr(\overline{\cM}_t) &=& \Pr\left( \target(\cT_{h^{\ast}_t}) < \log\frac{KL}{\delta} \cap  \Pr(\cT_{h^{\ast}_t}) < \frac{\log\frac{KL}{\delta}}{t_p} \right) \\
&\le& \Pr\left( \target(\cT_{h^{\ast}_t}) <\frac{\log\left(\frac{KL}{\delta}\right)}{\tau}  \cap  \frac{\target(\cT_{h^{\ast}_t})}{\rho_{h^{\ast}_t}(\source,\target)}< \frac{\log\left(\frac{KL}{\delta}\right)}{t_p} \right) \\
&\le& \Pr\left(\target(\cT_{h^{\ast}_t}) \le \min\left\{\frac{\log\left(\frac{KL}{\delta}\right)}{\tau}, \frac{\log\left(\frac{KL}{\delta}\right)}{h^{\ast}_t t_p}\right\}\right) 
\end{eqnarray*} 
Observe that this term is order-wise smaller than the per-step regret derived before. Using an integral approximation as before, we get that the regret is of the right order.
\end{itemize}

\subsection{Proof of Theorem~\ref{thm:special-family}}
\begin{proof}
We instantiate our results for a special family of distributions as in Assumption~\ref{assmpt:family-dissimilarity} as follows. Considering tree-discretized family of distributions, we have, $V_h^{\gamma}\rho_{h}(\source,\target) \le 1$. Therefore, equations~\eqref{eqn:st-a1t-final} become: 
\begin{eqnarray*}
\mE\left[s^{\ast}_t\ind{0 \le A_{1t} \le s^{\ast}_t}\right]&\le& c_{9} \max\left\{\frac{V^{\gamma}_{h^{\ast}_t}K\log\frac{KL}{\delta}}{t_p},\frac{V^{\gamma}_{h^{\ast}_t}K\log\frac{KL}{\delta}}{\tau}\right\}^{\left(\frac{(\alpha+1)}{\alpha}\right)\left(\frac{(\beta+1)}{\beta}\right)} \left(\frac{K\log\frac{KL}{\delta}}{\tau}\right)^{\frac{1}{\alpha}} \\
&\le& c_{9} \max\left\{\frac{K\log\frac{KL}{\delta}}{t_p},\frac{K\log\frac{KL}{\delta}}{\tau}\right\}^{\left(\frac{(\alpha+1)}{\alpha}\right)\left(\frac{(\gamma(\beta+1)}{\beta^{2}}\right)} \left(\frac{K\log\frac{KL}{\delta}}{\tau}\right)^{\frac{1}{\alpha}}
\end{eqnarray*}
Other arguments go as is.
\end{proof}

\section{Proof of Theorem~\ref{thm:multiple-shift}}
\label{appendix:thm2}
Let $h_t^{\ast}$ denote the optimal regression level. For $h \in \mN$ let $n_t(h)$ denote the covariate count from a distribution $\source_j$. Then, using argument identical to those in Lemma~\ref{lem:oracle-level}, we have that 
\[
\psi_t(h^{\ast}_t) \le c_1 V^{\beta}_{h^{\ast}_t} + \sqrt{\frac{\log\left(\frac{KL}{\delta}\right)}{\max\{\tau \target(\cT_{h^{\ast}_t}), \mE[n^{P}_t(h^{\ast}_t)]\}}}
\]
Further,
\[
\mE[n^{P}_t(h^{\ast}_t)] = \sum_{j}\frac{n_j}{n_t^{P}(h)} \source_{j}\left(\cT_{h^{\ast}_t}\right)
\]
Following arguments identical to Theorem~\ref{thm:cs-regret}, we have that $d_t = \sqrt{\frac{1}{\sum_{j}\frac{n_j}{n^{P}_{t}(h^{\ast}_t)}\source(\cT_{h^{\ast}_t})}}$. The inequality analogous to~\eqref{eqn:dissimilar-p}, we have: 
\begin{eqnarray*}
\mE_{Q}\left[d^{2}_{1t}\right] = \mE_{\target}\left[
\frac{1}{\sum_{j} \frac{n_j}{n_t^{P}(h^{\ast}_t)}\source_{j}} 
\right] &=& \min_{j}\mE_{\target}\left[\frac{1}{\frac{n_j\source_j}{n^{P}_{t}(h^{\ast}_{t})}}\right] = \min_{j} \frac{1}{\frac{n_j}{n_t^{P}(h^{\ast}_t)}} \mE_{\target}\left[\frac{1}{\source_{j}}\right] = \min_{j}\left\{\frac{\rho\left(\source_j,\target\right)}{\frac{n_j}{n^{P}_{t}(h^{\ast}_t)}}\right\}\\
&\le& \frac{1}{\sum_{j}\frac{n_j}{n^{P}_{t}(h^{\ast}_t)\rho(\source_j,Q)}}
\end{eqnarray*}
Since $x \rightarrow \frac{1}{x}$ is convex, by Jensen's inequality: 
\[
\sum_{j}\frac{n_j}{n^{P}_{t}(h^{\ast}_t)\rho(\source_j,Q)} \ge \frac{1}{\rho\left(\sum_{j}\frac{n_j}{n^{P}_{t}} P_j, Q\right)} = \frac{1}{\rho(\tilde{P},Q)}
\]
where, $\tilde{P}$ denotes the mixture measure. This can now be substituted in the proof of Theorem~\ref{thm:cs-regret} and all other arguments go through as is to obtain stated guarantees.

\section{Numerical Experiments}
\label{appendix:numerical-experiments}

The action space is given by $\cA=[K], \ K \ge 2$ and the context space is given by $\cX=[0,1]$. For the first two experiments, we consider a transfer learning setup where we are interested in learning under distribution $Q$ after observing $t_p < T$ contexts from distribution $P$. The distribution $Q \sim \text{Uniform}[0,1]$ and distribution $P \sim (\nu+1)x^{\nu}, \ x \in [0,1]$ and the dissimilarity increases as $\nu$ increases. We consider a biobjective problem wherein the reward function $\mu_{k}(x) = [\mu^{(1)}_{k}(x), \mu^{(2)}_{k}(x)]$ for the arm $k$. Define $k_{1}(x) = \frac{5}{4(1-x)}$ and $k_{2}(x) = \frac{5}{(5-4x)}$. 
\begin{eqnarray*}
\mu^{(1)}_{k}(x) &=& \max\left\{0,\left(1-5\left(\frac{1}{k}-\frac{1}{k_{(1)}(x)}\right)\right)\right\}\\
\mu^{(2)}_{k}(x) &=& 
\begin{cases}
\max\left\{0,\left(1-5\left(\frac{1}{k_{2}(x)}-\frac{1}{k}\right)\right)\right\}, \ \text{if} \ k  > k_{2}(x)   \\ 
\max\left\{0,\frac{1}{4}\left(\frac{1}{k}-\frac{1}{k_{2}(x)}\right)\right\}, \ \text{if} \ k \le k_{2}(x) 
\end{cases}
\end{eqnarray*}
It can be verified that given a context $x$, the Pareto-optimal arms are those with indices that lie in $[\lfloor k^{\ast}_{1}(x)\rfloor, \ldots, \lceil k^{\ast}_{2}(x) \rceil]$. Each experiment is conducted 10 times and the shaded area denotes one standard deviation in outcome. There are no known benchmarks for our problem and therefore, we use a policy that uses a random arm as the benchmark. We set $\delta = \frac{1}{T}, \sigma =1, \alpha = 0.2$ (this choice reflects a hard instance from the perspective of margin criteria).

\textbf{Environment Setup:}
The experiments are done on a MacBook Air with an Apple M1 chip, 16 GB memory and 10 core CPU. All codes are written in Python3 using several open source packages. The running time for all experiments ranges from less than a minute to a few hours. The code is available at: \href{https://anonymous.4open.science/r/CovarShift-B17A}{this link}.


\subsection{Effect of change-point}
\begin{figure}[htbp]
    \centering  
    \begin{minipage}[b]{0.5\textwidth}
        \centering
        \includegraphics[width=\linewidth]{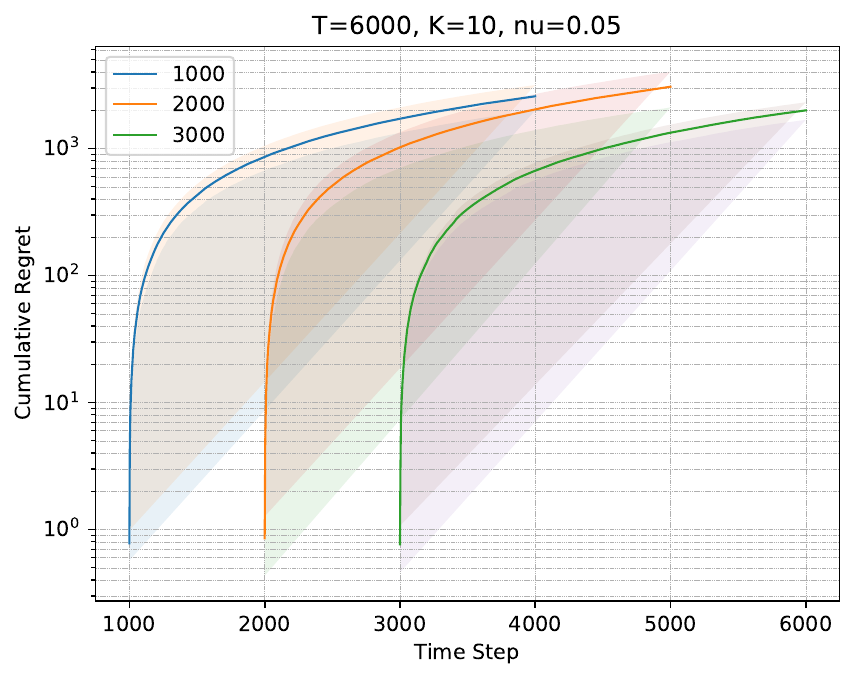}
        \caption*{}
    \end{minipage}\hfill
    \begin{minipage}[b]{0.5\textwidth}
        \centering
        \includegraphics[width=\linewidth]{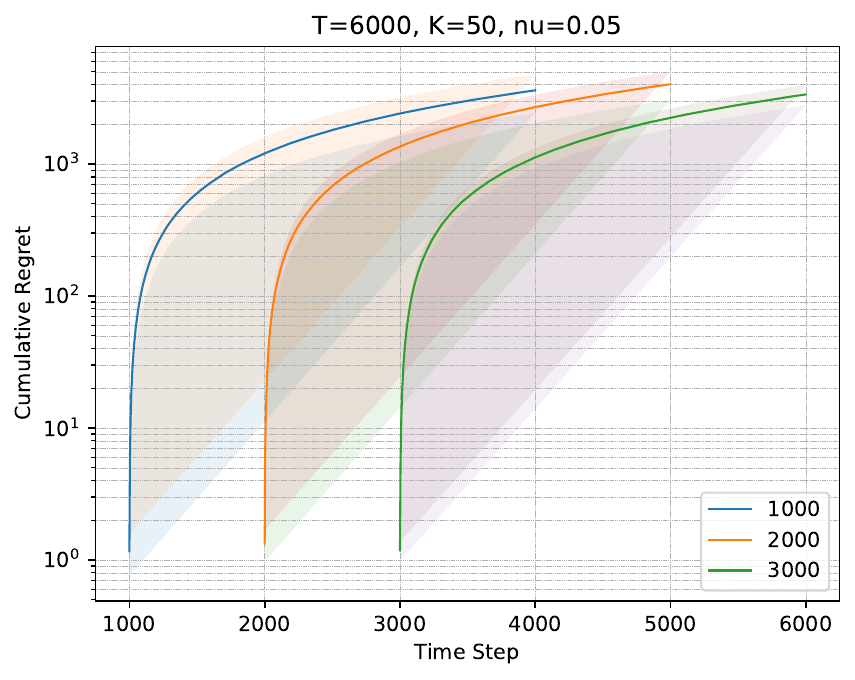}
        \caption*{}
    \end{minipage}
    \vspace{1em}

    \begin{minipage}[b]{0.5\textwidth}
        \centering
        \includegraphics[width=\linewidth]{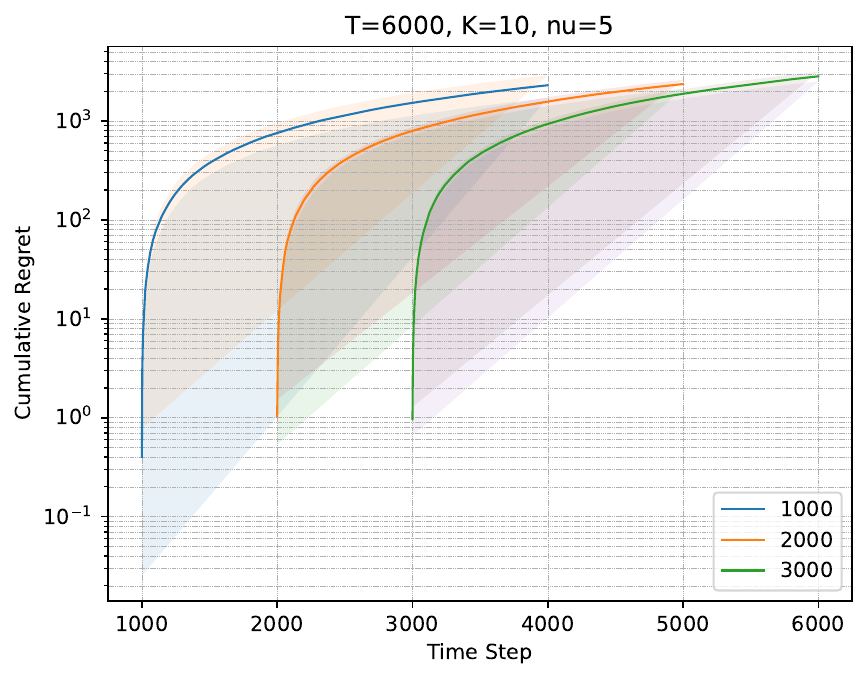}
        \caption*{}
    \end{minipage}\hfill
    \begin{minipage}[b]{0.5\textwidth}
        \centering
        \includegraphics[width=\linewidth]{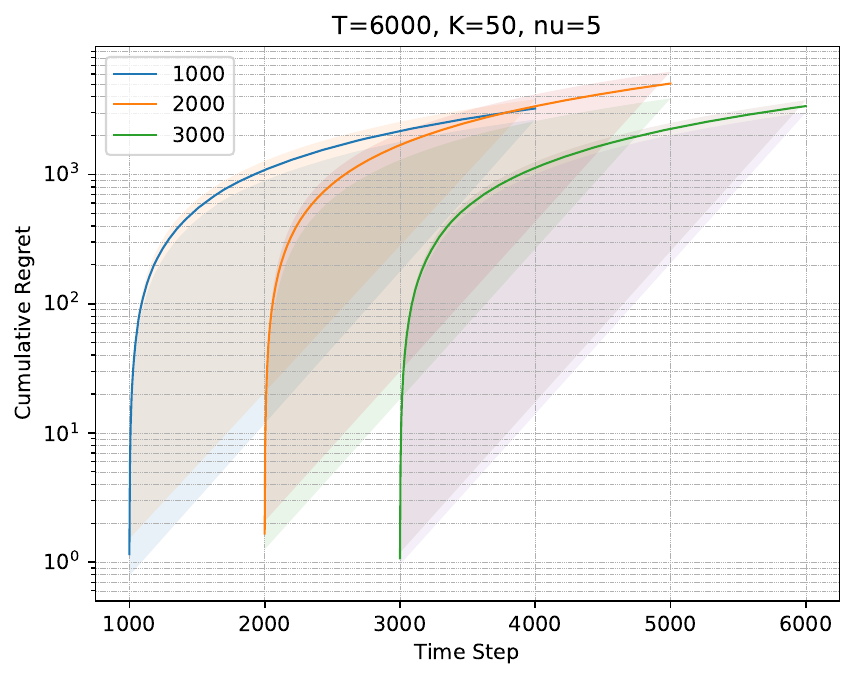}
        \caption*{}
    \end{minipage}
    \caption{Effect of changepoint $t_p$; the regret is measured for last $\tau = T-t_p$ samples.}
    \label{fig:effect-changepoint}
\end{figure}

We study the effect of regret over samples from target $Q$ by varying the number of samples drawn from source distribution $P$. We vary $t_{P}$ between $[1000,2000,3000]$ and keep $T=5e4$. Intuitively, as $t_p$ is increased the regret of the policy over $T-t_p$ rounds should decrease since more samples from source $P$ make it easier to learn under the target distribution $Q$. On the other hand, policies that allow for adversarial context arrival should display no change in performance since they are agnostic to learning about the target $Q$ using samples from source $P$. In Figure~\ref{fig:effect-changepoint}, we see this behavior as the number of arms and $\nu$ is varied. While the regret remains low when we have a larger number of samples from the source distribution, it worsens as the number of arms increases and $\nu$ is increased (Figures in the RIGHT panel have larger number of arms).

%

\section{Auxiliary Results}

\begin{lemma}[Generalization Result due to Vapnik and Chervonenkis, Theorem~$5.1$ in~\citep{bousquet2003introduction}] 
\label{lem:generalization-bound-vc}
Let $\cG$ be a function class from $\cX \rightarrow \{0,1\}$ with VC-dimension $d$ and $\source$ be a probability distribution on $\cX$. Let $\mE$ denote the expectation wrt $\source$ and $\mE_{n}$ denote the empirical expectation using iid samples from $\source$. Then for any $\delta > 0$, with probability $1-\delta$ it holds that for all $g \in \cG$: 
\[
-\min\left( \beta_n \sqrt{\mE_{n}(g)}, \beta^{2}_n + \beta_n\sqrt{\mE(g)} \right) \le \mE[g] -\mE_{n}[g] \le  
\min\left( \beta_n^{2} + \beta_n\sqrt{\mE_{n}[g]}, \beta_n\sqrt{\mE(g)} \right)
\]
where, $\beta_n = \sqrt{\frac{4}{n}(d\ln(2n)+\log(\frac{8}{\delta}))}$.
\end{lemma}

\begin{lemma}[Hoeffding's inequality]
Let \( X_1, X_2, \dots, X_n \) be independent random variables such that for all \( i \), the random variable \( X_i \) is bounded as: $a_i \leq X_i \leq b_i$ and $\bar{X}$ be the sample mean. Then, for any \( t > 0 \), Hoeffding’s inequality states:
\[
\mathbb{P} \left( \bar{X} - \mathbb{E}[\bar{X}] \geq t \right) \leq \exp \left( \frac{-2 n t^2}{\sum_{i=1}^{n} (b_i - a_i)^2} \right)
\]
and
\[
\mathbb{P} \left( |\bar{X} - \mathbb{E}[\bar{X}]| \geq t \right) \leq 2 \exp \left( \frac{-2 n t^2}{\sum_{i=1}^{n} (b_i - a_i)^2} \right)
\]
\end{lemma}
\end{document}